\documentclass[a4paper,fleqn]{cas-dc}

\usepackage[authoryear]{natbib}
\usepackage{graphicx}
\usepackage[inline]{enumitem}
\usepackage{url}
\graphicspath{{Figures/}}
\usepackage{caption}
\usepackage{hyperref}
\usepackage{pifont}
\usepackage{tikz}
\usepackage{tcolorbox}
\usepackage{geometry}
\usepackage{enumitem}
\usetikzlibrary{positioning, shapes.multipart}
\usetikzlibrary{positioning, arrows.meta}
\usepackage{siunitx}
\usepackage{subcaption}

\def\tsc#1{\csdef{#1}{\textsc{\lowercase{#1}}\xspace}}
\tsc{WGM}
\tsc{QE}
\tsc{EP}
\tsc{PMS}
\tsc{BEC}
\tsc{DE}
\newcommand{\xmark}{\ding{55}}

\begin{document}
\let\WriteBookmarks\relax
\def\floatpagepagefraction{1}
\def\textpagefraction{.001}
\shorttitle{Poultry Farm Intelligence}
\shortauthors{Panagi et~al.}

\title[mode = title]{Poultry Farm Intelligence: An Integrated Multi-Sensor AI Platform for Enhanced Welfare and Productivity}


\author[labelcyens]{Pieris Panagi}[orcid=0009-0007-4027-1488]\ead{p.panagi@cyens.org.cy}

\author[labelcyens]{Savvas Karatsiolis}[orcid=0000-0002-4034-7709]\ead{s.karatsiolis@cyens.org.cy}

\author[labelcyens]{Kyriacos Mosphilis}[orcid=0009-0004-3060-1568]\ead{k.mosphilis@cyens.org.cy}

\author[labelalgolysis]{Nicholas Hadjisavvas}[orcid=0009-0006-1426-9950]\ead{nicholas.hadjisavvas@algolysis.com}

\author[labelcyens,labeltwente]{Andreas Kamilaris}[orcid=0000-0002-8484-4256]\ead{a.kamilaris@cyens.org.cy}

\author[labelalgolysis]{Nicolas Nicolaou}[orcid=0000-0001-7540-784X]\ead{nicolas@algolysis.com}

\author[labelalgolysis]{Efstathios Stavrakis}[orcid=0000-0002-9213-7690]\ead{stathis@algolysis.com}

\author[labelcyens]{Vassilis Vassiliades\corref{cor1}}[orcid=0000-0002-1336-5629]\ead{v.vassiliades@cyens.org.cy}

\cortext[cor1]{Corresponding author}

\affiliation[labelcyens]{organization={CYENS - Centre of Excellence},
            addressline={Plateia Dimarchou Lellou Demetriadi 1}, 
            city={Nicosia},
            postcode={1016}, 
            country={Cyprus}}

\affiliation[labelalgolysis]{organization={Algolysis Ltd},
            addressline={Archbishop Makarios III 200}, 
            city={Lakatamia},
            postcode={2311}, 
            country={Cyprus}}

\affiliation[labeltwente]{organization={Department of Computer Science},
            addressline={University of Twente}, 
            city={Enschede},
            postcode={7522 NB}, 
            country={The Netherlands}}

\begin{abstract}
Poultry farming faces increasing pressure to meet productivity targets while ensuring animal welfare and environmental compliance. Yet many small and medium-sized farms lack affordable, integrated tools for continuous monitoring and decision-making, relying instead on manual, reactive inspections. This paper presents Poultry Farm Intelligence (PoultryFI) — a modular, cost-effective platform that integrates six AI-powered modules: Camera Placement Optimizer, Audio-Visual Monitoring, Analytics \& Alerting, Real-Time Egg Counting, Production \& Profitability Forecasting, and a Recommendation Module.

Camera layouts are first optimized offline using evolutionary algorithms for full poultry house coverage with minimal hardware. The Audio-Visual Monitoring module extracts welfare indicators from synchronized video, audio, and feeding data. Analytics \& Alerting produces daily summaries and real-time notifications, while Real-Time Egg Counting uses an edge vision model to automate production tracking. Forecasting models predict egg yield and feed consumption up to 10 days in advance, and the Recommendation Module integrates forecasts with weather data to guide environmental and operational adjustments.

This is among the first systems to combine low-cost sensing, edge analytics, and prescriptive AI to continuously monitor flocks, predict production, and optimize performance. Field trials demonstrate 100\% egg-count accuracy on Raspberry Pi 5, robust anomaly detection, and reliable short-term forecasting. PoultryFI bridges the gap between isolated pilot tools and scalable, farm-wide intelligence, empowering producers to proactively safeguard welfare and profitability.
\end{abstract}



\maketitle

\section{Introduction}

Advances in poultry farming continually raise the bar for both animal welfare and operational efficiency. Modern intensive systems demand constant monitoring of flock behavior, air quality, temperature, and production throughput, yet many small and medium-sized farms cannot afford the high initial investment or ongoing maintenance of traditional hardware and software solutions. As a result, farmers often rely on manual inspections and paper-based record keeping, which are time-consuming, prone to human error, and inherently lacking indicators of emerging problems. Regulatory bodies are increasingly mandating stricter welfare and environmental standards, while market pressures push for higher yields and lower unit costs, placing farm operators in a “double-bind” where under-resourced oversight undermines both compliance and profitability. Although recent research in sensor networks, edge computing, and AI-driven analytics has demonstrated isolated successes, such as audio-based disease detection or computer-vision egg counters, these tools typically operate in silos, forcing farms to cobble together point solutions rather than benefit from a unified, end-to-end platform.

To address these gaps, we introduce the Poultry Farm Intelligence (PoultryFI) system, a modular, low-cost platform that unifies six innovative modules, ranging from automated camera layout optimization to real-time egg counting, multi-modal welfare monitoring, predictive analytics, and decision support, into a single, end-to-end solution for modern poultry farm management.

Our solution prioritizes the overall welfare and productivity of the entire flock, rather than tracking individual birds with dedicated sensors. Individual monitoring is often invasive, error-prone, and labor-intensive, requiring significant effort to tag thousands of animals. In contrast, our approach is non-invasive, easy to deploy, and capable of delivering accurate, timely alerts to farmers, enabling effective and efficient flock management without disrupting the animals.

\begin{figure*}[ht]
  \centering
  \includegraphics[width=0.9\textwidth]{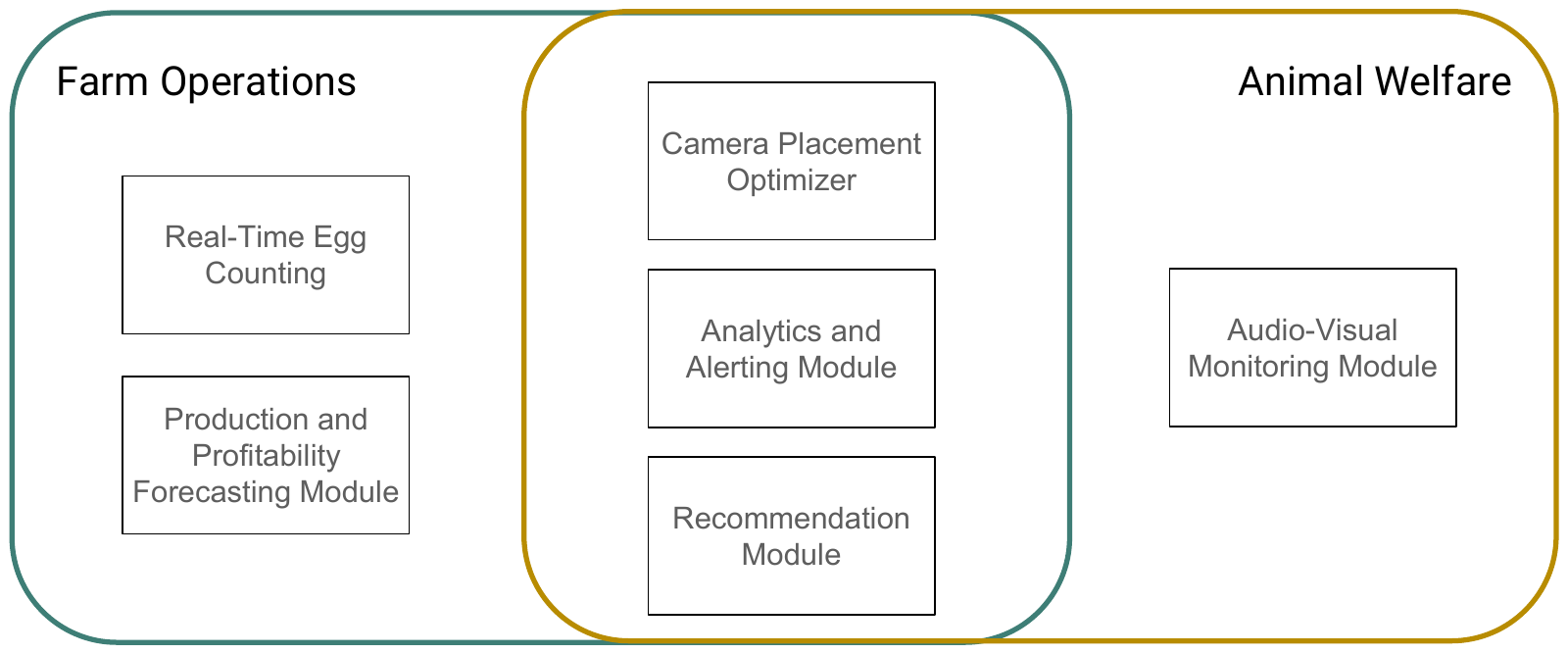}
  \caption{PoultryFI modules grouped by their primary impact on \textbf{Farm Operations}, \textbf{Animal Welfare}, or both. The \textbf{Real-Time Egg Counting} and \textbf{Production and Profitability Forecasting} modules are designed to streamline farm management and planning. The \textbf{Audio-Visual Monitoring Module} focuses specifically on monitoring animal welfare indicators. The \textbf{Camera Placement Optimizer}, \textbf{Analytics and Alerting Module}, and \textbf{Recommendation Module} contribute to both areas by transforming welfare-related data into actionable insights that support more efficient and responsive farm operations.}
  \label{fig:contributions}
\end{figure*}

\begin{enumerate}
    \item The ``Camera Placement Optimizer" automatically determines optimal camera positions and orientations by employing advanced evolutionary algorithms. This module achieves maximum surveillance coverage and delivers multiple high-quality configurations in a fraction of the time required by traditional, manually designed setups.
    \item The ``Audio-Visual Monitoring Module" harnesses synchronized audio and video streams from distributed sensor modules. On-device neural network inference is used to analyze livestock behavior and detect anomalies, providing real-time insights into animal welfare. 
    \item The ``Analytics and Alerting Module" aggregates data from audio, video, and environmental sensors to generate daily operational profiles, forecast trends, and issue proactive alerts when conditions deviate from established norms.
    \item The ``Real-Time Egg Counting Module" retrofits conventional egg grading machines with edge computer vision capabilities, ensuring accurate and real-time egg counting.
    \item The ``Production and Profitability Forecasting Module" integrates historical production data with sensor-derived analytics to predict flock productivity and compute essential economic metrics such as ``cost per egg".
    \item The ``Recommendation Module" combines internal analytics with external weather data to generate actionable advice aimed at optimizing farm operations.
\end{enumerate}
Together, these six modules form an integrated platform that advances poultry farm management towards improved animal welfare, operational efficiency, and overall profitability (see Figure~\ref{fig:contributions}).

 The main advantages of the proposed system are  
\begin{itemize}
\item Low-cost, robust hardware design and deployment
\item Continuous multi-modal data acquisition
\item Advanced unsupervised neural modeling for anomaly detection
\item Real-time inference with real potential for proactive intervention
\item A clear path to generalization and integration into commercial farms
\end{itemize}

The paper is organized as follows. In Section~\ref{sec-Background}, we present the Background, where we review key sensor technologies, AI and machine learning (ML) applications, optimal camera placement techniques, and the existing challenges and research gaps in poultry farm management. Section~\ref{sec-poultryfi-overview} provides an overview of the PoultryFI system, outlining its integrated framework and the overall architecture that combines all modules. Section~\ref{sec-description} describes each of the modules in detail and Section~\ref{sec-poultryfi-experiments} presents the experiments and evaluation of each module.  In Section~\ref{sec-discussion}, we provide a discussion on the lessons learned, the limitations, the scalability of our system, and outline future directions, and finally, Section~\ref{sec-conclusion} concludes the paper.

\section{Background and Related Work} \label{sec-Background}
\subsection{Overview of Precision Livestock Farming}

Precision Livestock Farming (PLF) represents a modern approach in animal agriculture that leverages Internet of Things (IoT) devices and AI/ML techniques to enable real-time monitoring and management of animal health, welfare, and productivity, addressing the challenges faced by both large-scale and small to medium-scale farms \citep{berckmans2017general, garcia2020systematic}. PLF employs diverse sensor modalities—such as visible light cameras, microphones, and environmental sensors—to collect rich data sets that inform applications ranging from behavior analysis and disease prediction to production optimization~\citep{ojo2022internet}. Economic constraints and stringent regulatory requirements further drive the need for cost-effective and scalable PLF solutions that can operate under varying conditions while ensuring compliance with established animal welfare guidelines. Despite significant advances in individual monitoring techniques, there remains a notable gap in the development of integrated, multi-modal systems that combine these technologies into a seamless platform, a gap that the PoultryFI system aims to bridge.

\subsection{Sensor Technologies and Data Acquisition in Poultry Farms}

Modern poultry farms rely on sensor technologies~\citep{astill2020smart} and data acquisition systems to monitor key parameters such as animal behavior, environmental conditions, and production dynamics in real time. A wide array of sensors—including visible light and infrared cameras, microphones, wearables, and environmental sensors that measure temperature, humidity, and gas concentrations—are used to capture diverse streams of data that support precise welfare and performance assessments~\citep{olejnik2022emerging,ojo2022internet}. These sensors are often integrated into compact, cost-effective devices capable of operating on edge platforms, thereby facilitating continuous, non-invasive monitoring even on small- and medium-scale farms~\citep{feiyang2016monitoring, kuccuktopcu2021comparison}. The rich, heterogeneous data collected necessitate efficient fusion and preprocessing techniques to generate actionable insights, ultimately forming the backbone of modern precision livestock farming strategies in poultry management~\citep{astill2020smart}.

\subsection{AI and ML Applications for Poultry Farm Management}
\subsubsection{Computer Vision and Sensor Placement Optimization}

\paragraph{Computer Vision:}
Cameras have become essential tools in modern poultry farming. By leveraging various types of cameras farmers and researchers can collect detailed visual data that supports automated systems for behavior analysis, disease detection, and individual tracking. These technologies help improve animal welfare, optimize resource use, and detect issues early, ultimately leading to more efficient and humane poultry management practices.

To detect broilers, \cite{guo2022monitoring} developed a deep learning approach supported by a custom behavioral dataset. They enhanced model performance through image augmentation techniques that accounted for varying poses and viewing angles of the birds, thereby capturing a broader range of behavioral features. In another study, \cite{cao2021automated} focused on using Convolutional Neural Networks (CNNs) to estimate chicken density by analyzing behaviors such as feeding, drinking, and jumping, achieving an accuracy of 93.8\%.

\cite{neethirajan2022chicktrack} introduced a computer vision system called ChickTrack, built on the YOLO framework, capable of identifying individual birds by detecting subtle physical differences. This capability enables persistent identification even when birds temporarily leave the camera's view or are obscured by poor lighting. The system can track individual birds and count them accurately, avoiding duplication. It also supports the detection of abnormal behaviors in the flock, such as changes in feed intake, feather pecking, improper lighting, or feed access. The incorporation of thermal imaging facilitates temperature monitoring at the individual level, which could aid in early illness detection. ChickTrack also enables tracking of movement patterns and prediction of unusual behaviors. Its mapping feature visualizes locomotion, gait, and trajectories, allowing for real-time issue detection. Additionally, it supports the study of how chickens respond to environmental changes without the influence of human presence.

According to \cite{ojo2022internet}, 2D visible light cameras have been used to assess posture and detect signs of illness or lameness in poultry. Thermal and infrared cameras provide additional functionality by capturing behavioral patterns, measuring body temperature, evaluating space usage, and monitoring stretching or laying behaviors. 3D cameras, such as Kinect devices used in \citep{okinda2019machine}, have been employed to observe and classify poultry health conditions, including the presence of Newcastle disease, using models like SVMs, ANNs, and logistic regression. Furthermore, \cite{nasiri2022pose} utilized Intel RealSense D455 to assess poultry posture and lameness. They adopted DeepLabCut~\citep{mathis2018deeplabcut}, a CNN-based pose estimation tool, in combination with an Long Short-Term Memory (LSTM) classifier to categorize mobility levels into four distinct groups.

\paragraph{Sensor Placement Optimization:}
Effective placement of cameras and sensors is essential in precision agriculture and livestock monitoring, as it enhances data quality, improves coverage, reduces blind spots, and ensures the efficient operation of automated systems. Properly positioned sensors contribute significantly to the accuracy of behavior tracking, environmental monitoring, and health diagnostics.

The following works have investigated strategies for optimizing the placement of cameras and other sensors. For instance, one study proposed a novel Field of View (FOV) model for visual sensor placement that mimics the distribution of the human eye to enhance monitoring effectiveness.

\cite{watras2018optimal} focused on determining the best camera poses for video stitching using a constrained greedy heuristic algorithm, while \cite{malhotra2022optimizing} explored optimal camera positions for achieving overlapping coverage through 3D camera projections.

Beyond visual monitoring, optimal sensor placements have also been studied in various domains such as agriculture~\citep{lee2019optimal} and agro-hydrological systems~\citep{sahoo2019optimal}, as well as in high-stakes environments like nuclear reactors, where constrained optimization techniques were applied to support the development of digital twins~\citep{karnik2024constained}.

In livestock-specific applications, \cite{sourav2022visual} presented an approach for optimal camera placement for bovine health monitoring using computer vision and Genetic Algorithms. Additionally, \cite{heyns2021optimization} utilized a multi-objective genetic algorithm, specifically the Non-dominated Sorting Genetic Algorithm II (NSGA-II)~\cite{deb2002fast}, to optimize camera localization for surveillance systems.

\begin{table*}[h]
    \centering
    \caption{Comparison between different existing systems and our own}
    \label{tab:system_comparison}
    \begin{tabular}{cccccc}
        \toprule
        \textbf{System} & \begin{tabular}[c]{@{}c@{}}Environmental\\ Sensors\end{tabular}
                       & \begin{tabular}[c]{@{}c@{}}Production\\ Data\end{tabular}
                       & \begin{tabular}[c]{@{}c@{}}Audio/Visual\\ Data\end{tabular}
                       & \begin{tabular}[c]{@{}c@{}}Productivity \\Forecasting\end{tabular}
                       & \begin{tabular}[c]{@{}c@{}}Tested on\\ Farm\end{tabular} \\
        \midrule
        \cite{sasirekha2023smart} & \checkmark & \xmark & \xmark & \xmark & \xmark \\
        \cite{karun2024smart} & \checkmark & \xmark & \xmark & \xmark & \xmark \\
        \cite{neethirajan2022chicktrack} & \xmark & \xmark & \checkmark & \xmark & \checkmark \\
        \cite{bumanis2023hen} & \checkmark & \checkmark & \xmark & \checkmark & \checkmark \\
        \cite{ji2025predicting} & \checkmark & \checkmark & \xmark & \checkmark & \checkmark \\
        \textbf{PoultryFI (ours)} & \checkmark & \checkmark & \checkmark & \checkmark & \checkmark \\
        \bottomrule
    \end{tabular}
\end{table*}

\subsubsection{Audio Monitoring and Welfare Assessment}

Microphones, also play a crucial role in poultry welfare assessment by enabling the non-invasive detection of vocal indicators related to health, stress, and behavior. Acoustic monitoring complements visual data and allows for continuous surveillance in large flocks, making it a valuable tool for early disease detection and environmental management. In recent years, deep learning (DL) techniques have been effectively applied to analyze audio data for improving poultry health monitoring systems.

Several studies have demonstrated the use of microphones in conjunction with DL methods for poultry welfare assessment. For example, \cite{cuan2020detection} employed Recurrent Neural Networks (RNNs) to detect respiratory issues in chickens using audio recordings, achieving a diagnostic accuracy of 97.4\%.
In another study, \cite{carpantier2019development} developed a technique capable of identifying sneezing sounds in poultry with 88.40\% precision, even in the presence of background noise.
Finally, \cite{ojo2022internet} reviews various approaches that use vocalizations to detect respiratory illnesses in poultry. Other methods have also been proposed to assess stress levels, feeding behavior, and growth rates based on the analysis of poultry vocal patterns.

\subsubsection{Forecasting, Analytics, and Edge-based Decision Support}
Accurate forecasting in poultry farming, particularly for egg production, plays a key role in optimizing operational planning and resource allocation. Recent research has explored data-driven approaches to improve prediction accuracy using both historical records and environmental sensor inputs. For instance, \cite{bumanis2023hen} evaluated various regression models to forecast egg production, achieving a Mean Absolute Percentage Error (MAPE) of 0.95\% using Random Forests. Similarly, \cite{ji2025predicting} applied environmental and production data to predict egg output in a broiler breeder farm, reporting a 1.82\% mean absolute error with an XGBoost model.

\subsection{Existing Systems}
Several systems have been developed that leverage IoT devices and Artificial Intelligence to streamline poultry farming operations and enhance productivity. However, most of these focus primarily on environmental sensor data, while others rely solely on vision-based modules without incorporating environmental measurements. Additionally, many of these systems have not been validated under real-world farm conditions. In contrast, our system combines low-cost audio-visual sensors with environmental sensors and production data provided by farmers to deliver welfare monitoring, forecasting, and timely alerts. A detailed comparison between existing solutions and our proposed system is presented in Table \ref{tab:system_comparison}.

\subsection{Challenges, Research Gaps, and Implications for PoultryFI}
Developing a comprehensive, low-cost solution for poultry welfare and farm efficiency remains a significant challenge. Most existing studies validate their proposed approaches in controlled environments, either in laboratory settings or experimental farms, which limits their real-world applicability. Moreover, identifying singular, reliable metrics that accurately capture the overall welfare of a flock continues to be a challenge.

In contrast, our work offers a holistic solution for monitoring animal welfare by introducing novel welfare indicators derived from visual and audio data. Simultaneously, we address key operational needs by incorporating predictive models for both environmental conditions within the poultry house and overall farm productivity. Finally, we offer a practical alert and Recommendation Module, designed to support informed decision-making by farmers. Notably, all modules have been deployed and validated in a commercial egg production setting, demonstrating the system’s applicability in real farming conditions.

\section{The PoultryFI System: An Overview}\label{sec-poultryfi-overview}

\begin{figure*}[ht]
  \centering
  \includegraphics[width=0.9\textwidth]{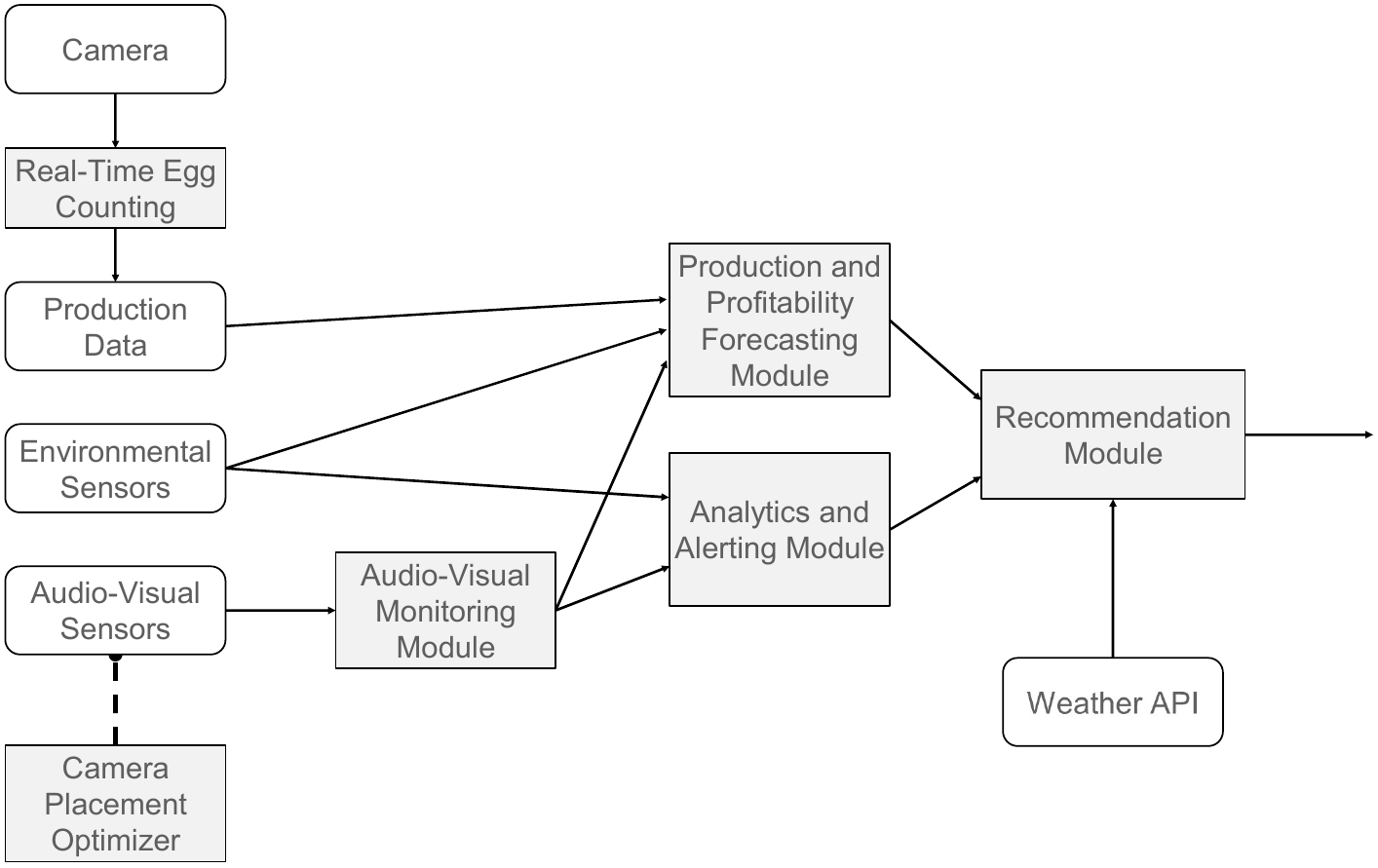}
  \caption{Overview of the PoultryFI System. Grey rectangles indicate the six core modules; white, rounded rectangles represent data sources. Solid arrows show the flow of information, while the dotted arrow indicates an offline process. The Camera Placement Optimizer determines optimal camera positions to maximize poultry house coverage and is executed offline. The Real-Time Egg Counting module supports egg counting by analyzing video data when farmers log production. The Audio-Visual Monitoring Module (AVMM) processes synchronized video and audio streams to generate welfare indicators, including motion score, audio anomaly score, and feeder status. The Analytics and Alerting Module (AAM) combines data from AVMM and environmental sensors (e.g., temperature, humidity) to issue alerts about short-term deviations in flock activity or environmental conditions. The Production and Profitability Forecasting Module (PPFM) uses production data, environmental data, and AVMM indicators to generate longer-term forecasts on poultry house productivity and profitability. Finally, the Recommendation Module synthesizes alerts from AAM, forecasts from PPFM, and external weather data to provide actionable recommendations for farm management.}
  \label{fig:diagram}
\end{figure*}

PoultryFI comprises six interlinked modules (Figure~\ref{fig:diagram}) that collectively enhance \textit{operational performance} and \textit{animal welfare}. Our system's approach is to move away from individual animal tracking and toward population-level behavioral insights, thus offering scalability and usability in real-world conditions. Moreover, the integration of motion and vocal cues adds redundancy and robustness to welfare assessments. Sensor and production data flow through these components to support tasks from optimal camera deployment to prescriptive decision support. Detailed methodology and evaluation are presented in Section~\ref{sec-description} and in Section~\ref{sec-poultryfi-experiments} respectively.

\begin{enumerate}
\item \textbf{Camera Placement Optimizer}: 
Identifying the optimal camera placement to ensure full visual coverage of the poultry house with as few devices as possible is crucial for the effective operation of our Audio/Visual Module with minimal cost. Manual attempts to address this challenge often result in suboptimal configurations with blind spots. To overcome this, the PoultryFI system includes an automated Camera Placement Optimizer that employs advanced evolutionary algorithms to generate high-quality camera layouts tailored to the specific dimensions and layout of each poultry house.

The module models the poultry house as a 2D area with discrete mounting beams and represents each camera by its position, orientation, depth, and field of view. Using evolutionary algorithms, it generates camera configurations that maximize spatial coverage, eliminating blind spots for both surveillance and behavioral observation. This optimization is performed prior to installation, ensuring that the deployed camera network achieves the desired coverage. By guaranteeing full visual coverage with fewer cameras, this module enhances operational monitoring efficiency and enables comprehensive welfare assessment.

\item \textbf{Audio-Visual Monitoring Module} (AVMM): 
Healthy, unstressed hens lay more eggs, yet subtle signs of distress or disease are difficult to detect across large flocks—manual checks are time-consuming and reactive. The AVMM overcomes this by collecting synchronized audio and video via distributed sensor nodes and extracting three key welfare indicators: bird activity through background-subtraction motion analysis, sound anomalies via an unsupervised autoencoder on Mel-spectrogram features, and feeder operation status using a lightweight classifier. These non-invasive metrics serve as early proxies for flock health and behavior, enabling prompt detection of stress or illness and directly advancing animal welfare.

\item \textbf{Analytics and Alerting Module} (AAM): Providing timely alerts is crucial for enabling farmers to take prompt action, helping prevent drops in productivity or flock welfare. To support this, our system includes an Alerting Module that delivers real-time alerts to the farmers.

This module aggregates AVMM indicators and environmental sensor readings (temperature, humidity) to compute daily statistical profiles. It issues real-time alerts when observed or short-term forecasted values deviate beyond learned thresholds. By flagging emerging welfare risks, AAM supports rapid operational interventions and proactive animal care.

\item \textbf{Real-Time Egg Counting}: Many small to medium-sized farms rely on standard mechanical machinery for egg grading and manual counting. To automate this process without requiring changes to existing equipment, our system includes a Real-Time Egg Counting Module that operates on an external Raspberry Pi 5.

The Real-Time Egg Counting Module applies an edge-deployed computer-vision model to video of existing egg-grading equipment in operation. It monitors the machine, detects and counts eggs in real time, replacing manual counting with automated, timestamped records that feed directly into production logs. This automation aims to reduce counting errors and labour, thereby enhancing farm productivity and data accuracy.

\item \textbf{Production and Profitability Forecasting Module} (PPFM): 
Flock productivity is a key indicator of a farm's profitability. Having the ability to estimate productivity levels in advance, allows farmers to make informed decisions and plan accordingly to maximize returns. To support this, our system includes a Production Forecasting Module that estimates flock productivity over a 10-day horizon.

Integrates historical egg counts, feed consumption, environmental conditions, and AVMM welfare indicators into time-series models that forecast egg output and feed requirements up to ten days ahead. It also calculates projected cost-per-egg. These data-driven forecasts enable more accurate budgeting, feed procurement, and labor planning, which are key to efficient farm management.

\item \textbf{Recommendation Module}: Combining the functionality of the modules described above to deliver practical recommendations is crucial for maximizing the overall effectiveness of the PoultryFI system.

The Recommendation Module receives AAM alerts, PPFM forecasts, and weather forecasts to generate prescriptive guidance. By linking operational metrics with welfare indicators, this module aims to deliver holistic recommendations that improve productivity while maintaining an environment conducive to flock health.
\end{enumerate}

The following section provides a detailed description of the methods and implementation of each module.

\section{Methods}\label{sec-description}

\subsection{Camera Placement Optimizer}
Strategic camera placement plays a critical role in achieving comprehensive visual coverage of the poultry farm, yet manual configuration is time-consuming, suboptimal, and difficult to adapt to different farm layouts. To address this challenge, the PoultryFI camera placement optimizer module uses advanced evolutionary algorithms to automatically generate high-quality camera configurations that are tailored to the spatial and structural characteristics of each farm. More specifically, we used Covariance Matrix Adaptation - Evolution Strategies~\citep[CMA-ES;][]{hansen2001completely} which is a state-of-the-art evolution strategy that returns a single globally optimal solution, and MAP-Elites~\citep{mouret2015illuminating,vassiliades2018discovering} which is a Quality Diversity optimization algorithm~\citep{chatzilygeroudis2021quality} able to return a diverse set of locally optimal solutions in a single run (by specifying the axes of diversity, i.e., behaviour descriptor). Below we provide a brief description of this module, discussing its most important aspects and technical details. For a more detailed description, we refer the reader to \cite{Mosphilis2025} and our open-source implementation\footnote{\url{https://github.com/CYENS/FarmCameraPlacementOptimizer}}.

\subsubsection{Formulation}
The optimizer aims to maximize coverage by $c \in \mathbb{N}_1$ cameras in a 2D rectangular farm of size $length \times width$. Cameras are mounted on predefined horizontal and vertical beams aligned with the X and Y axes, respectively.

Each camera has a capture depth $d_i$ and a horizontal field of view $\theta_i \in [0^\circ, 360^\circ]$. Coverage is represented as a binary image, where black pixels indicate monitored areas. Resolution is set via a user-defined pixel area.

\subsubsection{Genotype and Phenotype Representation}
Each camera is encoded as a chromosome with three genes: horizontal position, vertical position, and orientation. Gene values are normalized to $[0, 1]$ and scaled to real-world coordinates based on farm dimensions, with orientation mapped to the $0^\circ$–$360^\circ$ range. The orientation defines the direction the camera faces, determining the center of its field of view. For example, a camera oriented at $0^\circ$ with a $100^\circ$ field of view spans from $310^\circ$ to $50^\circ$.

\subsubsection{Fitness Function}
The fitness function measures how effectively the cameras monitor the farm by calculating the percentage of area covered. Each camera's field of view is modeled as an isosceles triangle, defined by its position, orientation, depth, and horizontal viewing angle. The triangle is rotated to match the camera's facing direction and projected onto a 2D grid representing the farm layout.

Coverage is computed by marking pixels within each triangle as monitored and calculating the ratio of monitored (black) pixels to the total number of pixels in the grid. The goal is to maximize this ratio, ensuring minimal blind spots across the farm.

\subsection{Audio-Visual Monitoring System for Livestock Welfare} \label{sec:audiovisual}

Ensuring chicken welfare on commercial farms is an inherently complex task. However, AI technologies can facilitate real-time data-driven  decision-making. In this context, we created a unified, cost-efficient monitoring system that collects environmental and behavioral data from audio and video cues using off-the-shelf components and then we process this data through an AI pipeline to infer analytics. In this section, we provide a brief description of the system and discuss some of its most important aspects and technical details. For a more detailed description, we refer the reader to~\citep{karatsiolis2024towards}.

\subsubsection{Implementation}
The system architecture is modular, comprising multiple sensor nodes and a central Synchronization and Processing Engine (SPE). Each sensor node is built on a Raspberry Pi 4B and includes a microphone, camera, USB speaker, and local SSD storage. These nodes are coordinated and managed by the SPE, built on a Raspberry Pi 5, which functions both as a Network Attached Storage (NAS) and as a computational node running the AI models for inference. 

A critical aspect of the system is the centralization of synchronization, processing, and logging. The SPE distributes synchronization messages and external stimuli (e.g., pre-recorded sounds for behavioral response tests), collects error logs, and coordinates audio-video acquisition across nodes. Moreover, it hosts AI models and runs motion detection and audio anomaly inference engines on data retrieved from the sensor modules. All hardware is deployed in custom 3D-printed protective enclosures to withstand the harsh farm environment characterized by dust, temperature fluctuations, and humidity. The data acquisition operates on a near-continuous basis: audio is recorded 24/7 in 58-second windows with a 2-second processing interval, while video is captured for 20 hours daily due to the limitation of visible-spectrum cameras (potentially upgradable to near-infrared in the future if required).

Indicative prices for the sensor nodes and NAS are shown in Tables \ref{tab:sensor-node-costs} and \ref{tab:nas-costs}. These prices cover the compute/storage stack (sensor nodes and NAS) and on-node peripherals (camera, microphone, and storage). Site-dependent items (e.g., environmental sensors, cables, mounts, and network gear) are excluded due to regional and farm-specific variability. The per-house total (compute/storage only) depends on the number of sensor nodes deployed.

\begin{table*}[t]
  \centering
  \caption{Sensor node components and indicative prices. Trade names are examples; equivalent products may be used.}
  \label{tab:sensor-node-costs}
  \begin{tabular}{@{} l l S[table-format=3.2] @{}}
    \toprule
    \textbf{Component} & \textbf{Specification (example)} & {\textbf{Indicative price [EUR]}} \\
    \midrule
    RPi 4B & 8 GB RAM & 140.00 \\
    RPi 4 Power Supply & 15 W & 12.00 \\
    Microphone & Fyvadio Lavalier & 11.00 \\
    Camera Module & RPi Camera Module 3 Wide & 61.00 \\
    Mini USB Stereo Speaker & 2 W & 7.00 \\
    480 GB SSD & Intenso & 50.00 \\
    32 GB SD Card & Intenso & 6.50 \\
    \midrule
    \multicolumn{2}{r}{\textbf{Total (per sensor node)}} & \bfseries 287.50 \\
    \bottomrule
  \end{tabular}
\end{table*}

\begin{table*}[t]
  \centering
  \caption{NAS / synchronization \& processing engine components and indicative prices. Trade names are examples; equivalent products may be used.}
  \label{tab:nas-costs}
  \begin{tabular}{@{} l l S[table-format=3.2] @{}}
    \toprule
    \textbf{Component} & \textbf{Specification (example)} & {\textbf{Indicative price [EUR]}} \\
    \midrule
    RPi 5 & 8 GB RAM & 90.00 \\
    RPi 5 Power Supply & 15 W & 10.00 \\
    16 TB HDD & NAS-grade (e.g., Seagate IronWolf Pro) & 250.00 \\
    Hard Drive Enclosure & ORICO & 27.00 \\
    32 GB SD Card & Intenso & 6.50 \\
    \midrule
    \multicolumn{2}{r}{\textbf{Total (NAS per house)}} & \bfseries 383.50 \\
    \bottomrule
  \end{tabular}
\end{table*}

\subsubsection{Audio-Based Anomaly Detection}
A central innovation of the system is its unsupervised audio anomaly detection model designed to infer the presence of alarming psychological states of chickens, such as stress and panic, based on their vocalizations. Unlike prior work that relies heavily on supervised learning and annotated datasets \citep{neethirajan2022chicktrack, 14,7}, we argue that annotating chicken emotional states is highly unreliable due to the difficulty to locate experts labor-intensive, and non-generalizable across farms. Instead, we adopt an unsupervised learning framework that leverages the farm's own historical audio recordings to construct a benchmark distribution of "normal" vocal behavior. The system splits each 58-second audio recording into 2-second chunks and converts these into Mel spectrograms, i.e., time-frequency representations that better align with human and animal auditory perception. Each spectrogram consists of an 80×80 matrix generated using 80 spectral filter banks and an FFT window size of 2220. Multiple Mel spectrograms from different sensor nodes are aggregated into a multichannel tensor. This technique allows the model to process sound events captured from different spatial positions simultaneously. This spatial fusion provides robustness against local occlusions or noise dominance at any single node. The core model is a Convolutional Denoising Autoencoder (Conv-DAE) with an encoder-decoder architecture. The encoder is based on the ResNet18 convolutional backbone and maps each spectrogram into a 512-dimensional latent space. The decoder consists of convolutional and upsampling layers that reconstruct the original spectrograms from their noisy versions. The autoencoder is trained using a Gaussian noise corruption process ($\sigma^2 = 0.25 $), with reconstruction loss measured by Mean Squared Error (MSE).

\subsubsection{Audio-Based Feeding Detection}

Missed or delayed feedings can quickly stress poultry and diminish productivity, yet continuously monitoring feeder mechanisms by sight or simplistic sound‐level thresholds is both impractical and unreliable. Instead, the distinct acoustic patterns produced by both the flock and the feeding machinery present a reliable signal for detecting feeder activity. Building on this observation, we implemented an audio‐based feeding detection module that leverages our existing anomaly detection encoder to distinguish between ``feeder open'' and ``feeder closed'' states in real time. 

\paragraph{Methodology:} Our methodology begins by passing each two‐second audio clip through the frozen encoder's bottleneck layer, extracting a 512‐dimensional feature vector that captures the essential characteristics of normal and feeder‐related sounds. We then train a lightweight feedforward neural network on these embeddings, using a custom‐labeled dataset in which clips are tagged according to whether the feeder was active. By freezing the encoder weights and tuning only the classifier, we can rapidly learn to map sound features to feeding events without degrading the encoder's general anomaly‐detection capabilities.

\subsubsection{Video-Based Motion Detection Methodology}
Complementary to audio analysis, the system includes a visual motion detection module. We use the Mixture of Gaussians Version 2 (MOG2) algorithm for background subtraction, which probabilistically models pixel distributions over time to distinguish between static and dynamic regions \citep{zivkovic2004improved}. MOG2 is ideal for handling illumination changes and shadows, both of which are common in farm environments. Each frame from the 30 fps video streams is processed through the MOG2 algorithm to yield binary motion masks. These masks indicate motion per pixel and are aggregated across frames to yield a motion score for each recording. This motion score becomes a proxy for the activity level of the flock, which can be correlated with welfare states. High activity might indicate panic or aggressive behavior, while consistently low activity could signal lethargy due to overfeeding or other causes~\citep{baybay2024lethargy}. Thus, this metric enables dynamic thresholds to be set for automated alerts, ranging from warning messages about possible predators to suggestions for environmental adjustments.

\subsection{Analytics and Alerting System}\label{sec:aas-description}
The environmental sensors integrated into our system provide valuable metrics for assessing the welfare of the flock. Temperature fluctuations, such as heat stress and cold stress, can have detrimental effects on both animal well-being and overall productivity~\citep{kim2024heat}. Therefore, the ability to predict significant rises or drops in temperature is critical, as it enables farmers to implement timely and appropriate precautionary measures.

Similarly, humidity levels play a key role in flock health. Low humidity is linked to an increased risk of respiratory issues~\citep{xiong2017humidity}, while high humidity impairs the birds' ability to regulate body temperature, thereby exacerbating thermal stress.

Furthermore, the audio/video indicators comprise information rich data, which when combined with environmental sensor data and production metrics (e.g., egg count, mortality rate) may provide temporal trend analysis of flock activity and stress markers, predictive analytics for disease outbreaks, root-cause identification of behavioral anomalies (e.g., high ammonia levels, temperature extremes) and recommendations for environmental control (ventilation, lighting, feeding).

\subsubsection{Forecasting Model}
To address these challenges, our system includes a short-term environmental forecasting model capable of predicting both temperature and humidity levels within the poultry house over a 3-hour time horizon. The model is based on Linear Regression and utilizes a combination of the current value of the target variable and its historical trends—specifically, the average values recorded over the past 3, 7, and 14 days—within a 3-hour look-back window.

The forecasting approach is iterative: the model initially generates a one-hour-ahead prediction, which is then used as an input to forecast the subsequent hour. This process is repeated until the full 3-hour forecast is produced.

\begin{figure}[ht]
    \centering
    \includegraphics[width=1\linewidth]{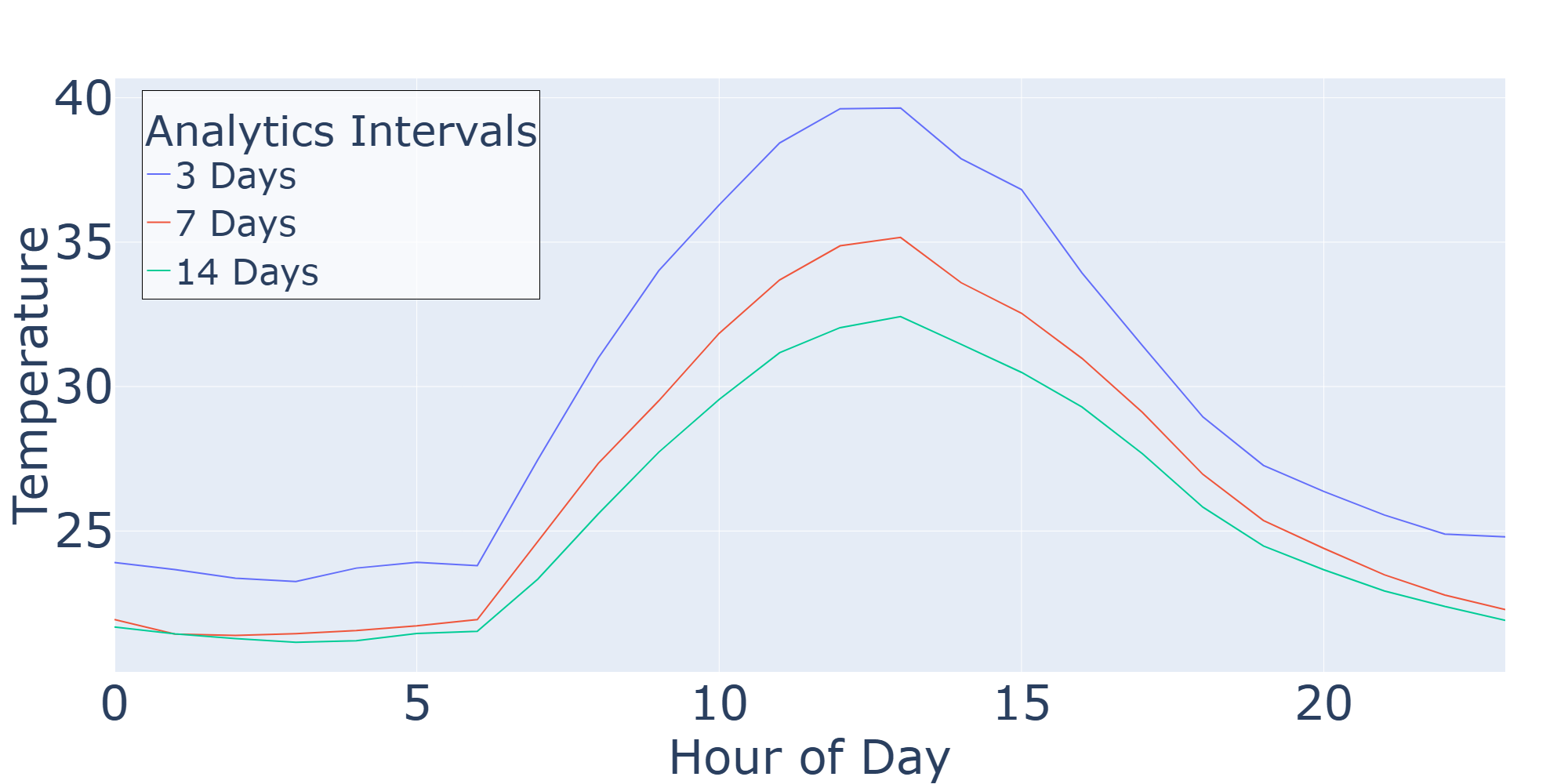}
    \caption{The profile of Temperature on 20/04/2025 for multiple day intervals}
    \label{fig:Temp_Profile}
\end{figure}

\begin{figure}[ht]
    \centering
    \includegraphics[width=1\linewidth]{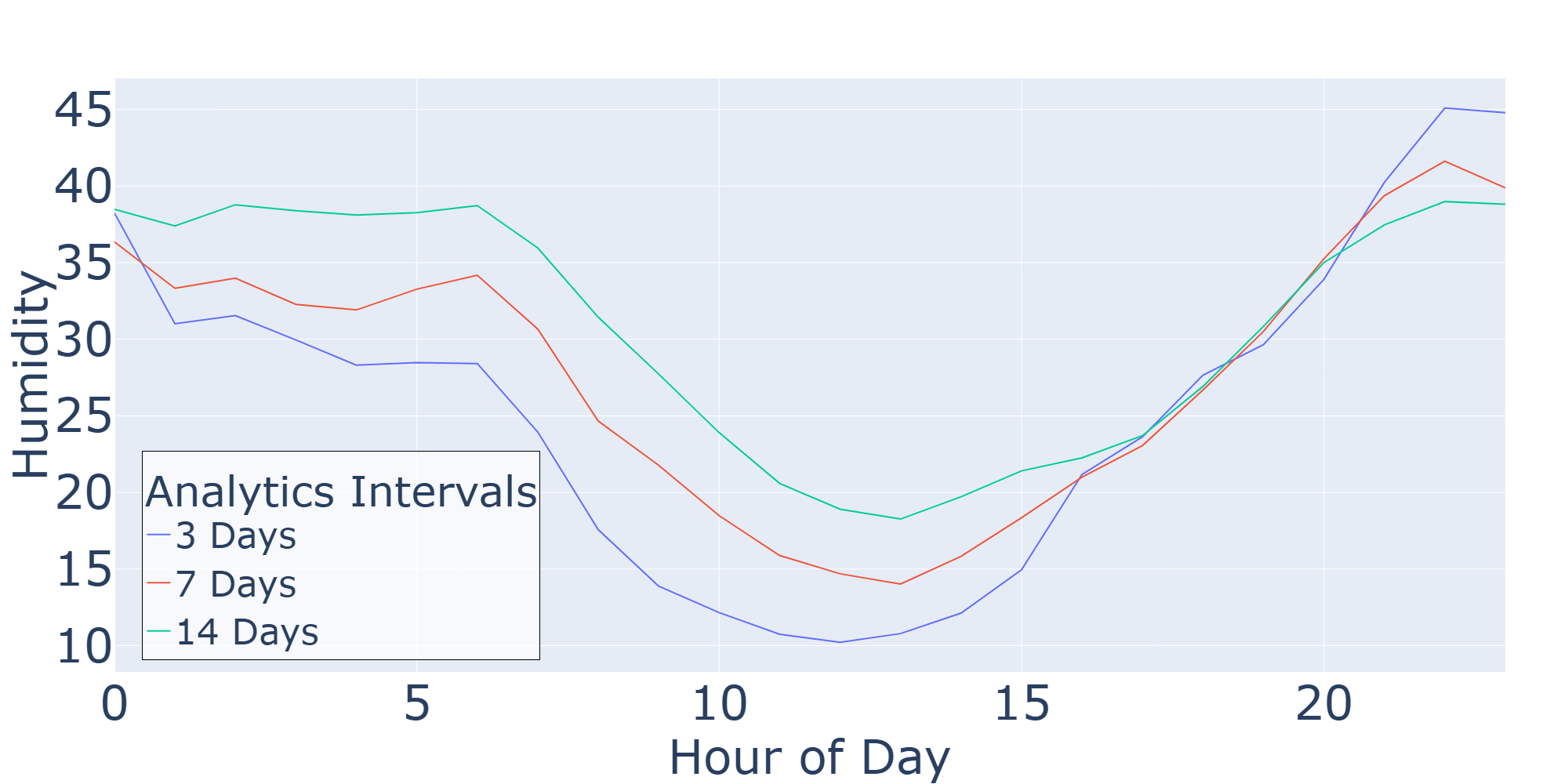}
    \caption{The profile of Humidity on 20/04/2025 for multiple day intervals}
    \label{fig:Hum_Profile}
\end{figure}

\subsubsection{Early Warning System}
Providing timely alerts to farmers about critical events on the farm is essential for ensuring flock welfare, enhancing cost-efficiency, and improving overall productivity by identifying and addressing potential operational bottlenecks. Our system generates alerts when significant increases or decreases in temperature and humidity are detected or forecasted, as outlined in the previous paragraph, thereby allowing farmers sufficient time to take corrective actions and prevent potentially adverse effects.

The audio and video indicators generated by the AI models, described in Section~\ref{sec:audiovisual} are used to issue alerts to farm owners when abnormal vocalizations or flock movements are detected. Specifically, the SPE runs an API service that sends alerts to a web server that maintains alert statistics and notifies farmers via email upon receiving an alert.

Finally, the system issues alerts based on insights from the audio and video analysis modules, as described in Section~\ref{sec:audiovisual}. Finally, it can also notify farmers of irregularities during feeding times by leveraging the feeding detection model.

\subsection{Real-Time Egg Counting}

The Real-Time Egg Counting module retrofits a standard Riva-Selegg grading machine with a ceiling-mounted Raspberry Pi 5 and Pi Camera V2, capturing top-down video of the conveyor’s feeding lane. A lightweight single-shot detector (EfficientDet-Lite0) is trained on a custom egg dataset and deployed on the Pi to localize each egg in 640×480px frames. Detections feed into a distance-based centroid tracker that assigns persistent IDs to eggs as they move through the field of view.

Counting occurs within four polygonal “bins” corresponding to the machine’s weight gates: extra-large, large, medium, and small. Each egg’s tracked centroid is tested against these regions via a point-in-polygon test; when an egg crosses into its designated bin it increments that class’s tally.
The calibration is automated; the operator runs at least twelve small eggs through the grader so they traverse the full lane. The detection software aggregates detected centroids, on which we apply DBSCAN to isolate high density clusters where springs momentarily halt the eggs. We then fit straight lines through the aggregated centroids with a Hough transform algorithm to recover the weighting springs positions. The positions belonging to the same weight class are merged to yield the regions of interest (ROIs) designated for egg detection. The different ROIs are combined to form the detection region outside of which all image pixels are masked, boosting inference precision and speed.

All components (image acquisition, inference, tracking, counting and calibration) run as a Python back end on the Pi; a small ReactJS front end lets operators adjust calibration, start/stop counting, and confirm tallies.

\subsection{Production and Profitability Forecasting System}
\subsubsection{Egg Production Forecasting}
As discussed in Section~\ref{sec:audiovisual}, the audio and video indicators reflect the vocalization intensity and motion index of the flock. In this context, they include (among other data) relevant information for predicting future egg production. These indicators are treated as cues for assessing the psychological state of the flock, as well as indirect evidence of potential stressors—such as starvation, dehydration, or excessive heat—that may lead to reduced egg production in the following days. In addition to the audio/video indicators computed by the AI pipeline, we also incorporate data recorded daily by the farmers, including total egg production and mortality. We further consider the age of the chickens (in weeks), as egg production naturally declines with age.

According to several studies~\citep{kim2015stress, shini2009stress}, the impact of a stressor on egg production—and on the overall welfare of the flock—may not become apparent until several days after the initial incident. This implies that a stress condition occurring today may not lead to a noticeable decline in egg productivity until up to a week later. In this context, we generate features based on various intervals, e.g., last 15 days, 7 days and 3 days and we predict the average egg production for the next 10 days.

In these lines, we developed a linear regression model that estimates the average egg production of the next 10 days. The model leverages multimodal data sources, including environmental sensor readings, visual and audio inputs from on-site cameras and microphones, as well as production data provided directly by farmers.We hand-engineered 40 features from the available data and identified the most influential ones through experimentation, that is, those that yielded the best predictive performance (see Section~\ref{sec:ppf-evaluation}). 

Additionally, we derive the predicted average productivity per bird by dividing the forecasted total egg production by the current number of chickens in the poultry house. To estimate the predicted cost per egg, we first developed a method for calculating the flock’s estimated daily food consumption. Farmers provided monthly records detailing the quantity and cost of food purchased for each poultry house. To compute the average daily food consumption, we summed the total amount of food acquired during the current production cycle—from its start until the most recent day—and divided this by the number of days the flock had been in the poultry house.

Using this value, we calculated the average daily food cost. We then estimated the predicted cost per egg over the next 10 days by dividing the calculated daily food cost by the predicted number of eggs produced during that period.

\subsection{Recommendation Module}
Our system is also equipped with a Recommendation Module, designed to support farmers in making informed decisions regarding their livestock. To generate actionable suggestions, the system aggregates and processes all collected and forecasted data—outlined in the previous sections—and uses this information as input to the Recommendation Module. Recommendations are generated upon request, providing farmers with context-aware guidance tailored to current and anticipated conditions within the poultry house.

Specifically, the Recommendation Module utilizes information related to potential alerts generated by the environmental sensor forecasting module. When such alerts are detected, the system provides targeted recommendations to the farmers. These suggested actions are based on relevant literature~\citep{li2025Recommendation} and expert knowledge gathered through consultations with field experts.

The Recommendation Module also takes into account potential alerts derived from audio and video indicators. When anomalies are detected, the system provides farmers with possible explanations for these irregularities, along with suggested actions to help restore the indicators to their expected levels for the corresponding hour of the day.

The system also considers the forecasted production values generated by the module described in Section~\ref{sec:aas-description} to provide farmers with relevant insights. This information supports decision-making regarding the optimal timing for flock culling, particularly when a decline in productivity is anticipated.

Finally, the current module incorporates data from a weather forecasting API (pypi.org/project/python-weather), which provides predictions for temperature, humidity, rain probability, and cloud coverage over the next three days in the area of the farm. This information is used to generate context-specific recommendations, such as advising farmers to test equipment in advance or implement precautionary measures that can help maintain flock welfare under the anticipated weather conditions.

\section{Experiments and Results}\label{sec-poultryfi-experiments}

All system modules described in this section were deployed and tested in a commercial egg farm located in Cyprus. The open poultry house where the deployment took place allowed for free movement of chickens without the use of cages and had a capacity of 1,200 chickens and, during the deployment period, housed approximately 750 chickens of a local breed. The structure measures 20.5 meters in length and 6.5 meters in width. Summers in the area of the farm are hot, dry, and sunny, with daytime highs often ranging from 30°C to 35°C, sometimes peaking above 40°C during heatwaves. Winters are mild and wetter, with daytime temperatures generally between 10°C and 16°C and occasional nighttime frosts. The poultry house is equipped with fans, heaters and water sprayers.

\subsection{Camera Placement Optimizer}

\subsubsection{Poultry house and Camera Specifications}
The poultry house measures $20.5$ meters in length and $6.5$ meters in width, with a resolution of $0.1 \times 0.1 m^2$ per pixel. A total of $6$ Raspberry Pi Camera Module $3$ devices are used, each with a $102^\circ$ field of view and a fixed depth of $5$ meters. $6$ camera modules are used as the total area that should be covered in the farm is $133.25m^2$ and each camera can cover approximately $\frac{102^\circ}{360^\circ}5^2\pi\approx22.25m^2$. The minimum number of cameras needed to cover the whole farm can be found by ceiling the ratio of farm area to the coverage area of a single camera: $\frac{133.25} {22.25}\approx\lceil{5.99}\rceil=6$. Cameras can be mounted on eight structural beams—three horizontal (at $0m$, $3.25m$, and $6.5m$) and five vertical (at $0m$, $5.125m$, $10.25m$, $15.375m$, and $20.5m$), with the outer beams marking the farm’s perimeter. The genotype encoding comprises 18 values in total, corresponding to three parameters per camera.

\subsubsection{Behaviour Descriptor}
To encourage solution diversity in MAP-Elites, we used a binary descriptor indicating beam utilization. Each dimension corresponds to a beam and has a value of $0$ (unused) or $1$ (used), forming a binary vector of length equal to the total number of beams. This descriptor enables the algorithm to explore combinations ($256$ in total) where different sets of beams are selected for camera placement.

\begin{figure}[ht]
    \centering
    \includegraphics[width=\linewidth]{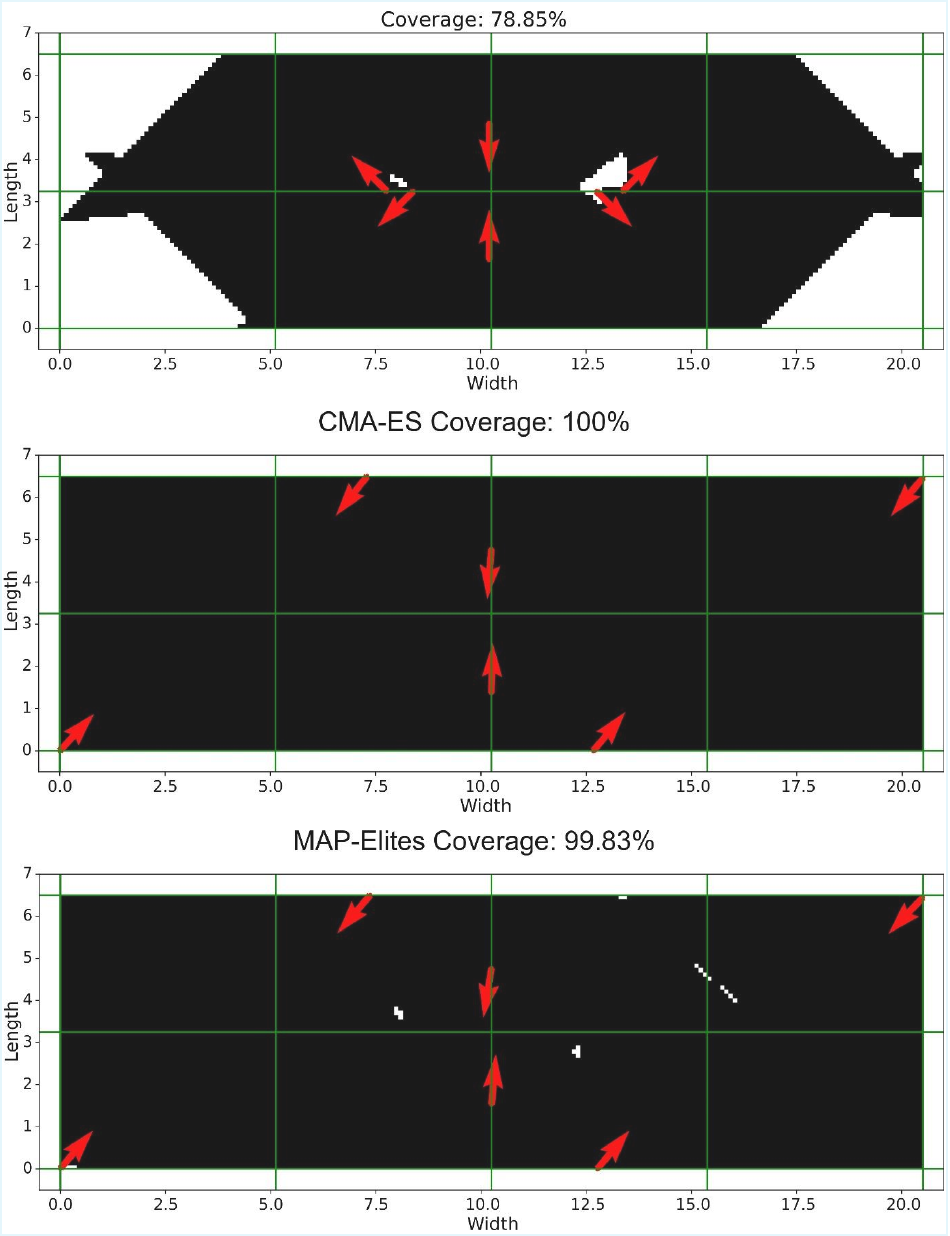}
    \caption{(Top) The Initial Manual Solution. (Middle) Solution provided by CMA-ES that was used to deploy the cameras in the poultry house. (Bottom) Solution provided by MAP-Elites similar to the deployed solution. Figures adapted from \cite{Mosphilis2025}.}\label{fig:solutions}
\end{figure}

\subsubsection{Results}

Both CMA-ES and MAP-Elites provided solutions achieving maximum coverage. More specifically, CMA-ES provided a solution with $100\%$ coverage which remarkably exhibited bilateral symmetry and was the one deployed in the farm (Figure~\ref{fig:solutions} Middle). MAP-Elites provided a range of unique solutions, one of which was almost identical to the solution provided by CMA-ES (Figure~\ref{fig:solutions} bottom). For more details, we refer the reader to \citep{Mosphilis2025}.

\subsection{Audio-Visual Monitoring System for Livestock Welfare} 

\subsubsection{Audio-Based Anomaly Detection}
Training achieves a reconstruction error of 0.126 on the training set and 0.141 on the test set, which indicates a reasonably well-fitted model. The model is trained on a dataset of 75,000 audio samples collected over 50 days from two sensor nodes.

Anomalies are detected based on high reconstruction errors. The underlying hypothesis is that audio patterns significantly deviating from the learned manifold of normal behavior will not be reconstructed well, thus producing a high error. To visualize and interpret these embeddings, we reduce the 512-D representations to 3-D via Principal Component Analysis (PCA) and observe spatial clustering of audio behaviors. Further, k-means clustering (k=5) is applied to the embeddings, revealing distinct semantic clusters: Flock resting (low-intensity sounds), Normal soft clucking, Calm clucking and ambient noises,  Machinery-induced noise and very soft vocalizations and Intense, chaotic sounds associated with panic or stress. Anomalous points (higher reconstruction error) predominantly fall within or near the same cluster, validating the model’s ability to isolate potentially alarming flock behaviors. These points are also linked to known triggers such as predator attacks or loud environmental disturbances. For more details, see \citep{karatsiolis2024towards}.

\begin{table*}[ht]
\centering
\caption{Audio-based feeding detection model configuration parameters and evaluation metrics. The best performing model (lowest validation/test loss and highest AUC score) is shown in bold.} \label{tab:feedingdetection}
\begin{tabular}{llccc}
\toprule
\multicolumn{2}{c}{\textbf{Model Configuration}} & \multicolumn{3}{c}{\textbf{Evaluation Metrics}} \\
\cmidrule(r){1-2} \cmidrule(l){3-5}
\textbf{Nodes per Hidden Layer} & \textbf{Weight Decay} & \textbf{Final Validation Loss} & \textbf{Test Loss} & \textbf{AUC Score (Test Set)} \\
\midrule
0 (logistic regression) & 0 & 0.155 & 0.155 & 0.9829 \\
0 (logistic regression) & 0.01 & 0.154 & 0.154 & 0.9835 \\
64 & 0 & 0.175 & 0.162 & 0.9914 \\
128 & 0 & 0.162 & 0.153 & 0.9927 \\
64 & 0.01 & 0.093 & 0.085 & 0.9936 \\
128 & 0.01 & 0.094 & 0.085 & 0.9936 \\
64, 64 & 0.01 & 0.097 & 0.084 & 0.9934 \\
128, 128 & 0.01 & 0.091 & 0.082 & 0.9943 \\
\cellcolor[gray]{0.9}\textbf{128, 32} & \cellcolor[gray]{0.9}\textbf{0.01} & \cellcolor[gray]{0.9}\textbf{0.088} & \cellcolor[gray]{0.9}\textbf{0.077} & \cellcolor[gray]{0.9}\textbf{0.9944} \\
\bottomrule
\end{tabular}
\end{table*}

\subsubsection{Audio-Based Feeding Detection}

\paragraph{Experimental Evaluation:} We evaluated performance using 15498 ``feeder open'' and 30996 ``feeder closed'' clips. Data were split into training (60\%), validation (20\%), and test sets (20\%). To mitigate class imbalance, each training epoch included all minority‐class samples plus a randomly undersampled subset of majority‐class clips of equal size. The maximum number of epochs was set to 200 and the batch size to 2048. We framed the task as binary classification, optimizing binary cross‐entropy loss and applying early stopping based on validation error (with an early stopping patience value of 70). We used the Adam optimizer \citep{kingma2015adam} with a learning rate of 0.001.
We compared three classifier architectures, i.e., with no hidden layer (equivalent to logistic regression), with one and two hidden layers of ReLU neurons by varying the number of nodes per layer, with and without L2 regularization, the latter expressed as a weight decay term in Adam optimizer with a value of 0.01. The fixed hyperparameters were chosen after preliminary experimentation.
Final performance was measured on the held‐out test set using both classification error and the area under the ROC curve (AUC), which captures the trade‐off between true and false positive rates across all decision thresholds. The results are shown in Table~\ref{tab:feedingdetection} where we observe that: (1) having a hidden layer results in better performance (lower validation loss and higher AUC score), (2) weight decay improves all models and the (high) value used prevents overfitting, and (3) the best performing model is one that uses two hidden layers of 128 and 32 neurons which was the one deployed on the SPE.

\paragraph{Deployment:} To monitor feeder activity in practice, we need to determine not only whether the feeder opened but also the duration it remained open. This requires reliably detecting the start and end times of each feeding event. To address this, we deployed the model within a script that processes a sequence of 27 consecutive 2-second audio clips. If most of these clips are classified as "feeder open", the system logs that the feeder is active; otherwise, it records it as closed. This majority-vote smoothing improves robustness to occasional misclassifications, which are expected in noisy real-world environments.

\begin{table*}[h]
    \centering
    \caption{The Root Mean Square Error (RMSE) for the test set for Temperature and Humidity for different configurations}
    \label{tab:short_term}
    \begin{tabular}{ccccccccccc}
        \toprule
        \multirow{3}{*}{\textbf{Model}} & \multicolumn{2}{c}{\textbf{Configuration}} & \multicolumn{8}{c}{\textbf{Test Set RMSE}} \\
        \cmidrule(lr){2-3} \cmidrule(lr){4-11}
        & \textbf{Historical Profile} & \textbf{Look-back Window} & \multicolumn{2}{c}{\textbf{2 Hours}} & \multicolumn{2}{c}{\textbf{3 Hours}} & \multicolumn{2}{c}{\textbf{4 Hours}} & \multicolumn{2}{c}{\textbf{5 Hours}} \\
        \cmidrule(lr){4-5} \cmidrule(lr){6-7} \cmidrule(lr){8-9} \cmidrule(lr){10-11}
        & & & \textbf{T} & \textbf{H} & \textbf{T} & \textbf{H} & \textbf{T} & \textbf{H} & \textbf{T} & \textbf{H} \\
        \midrule
        Linear Regression & \xmark  & 1 Hour & 1.03 & 1.93 & 1.98 & 1.49 & 4.24 & 3.20 & 5.95 & 6.27 \\
        Linear Regression & \xmark  & 2 Hours & 1.71 & 2.12 & 1.60 & 1.23 & 3.12 & 2.12 & 4.91 & 3.32 \\
        Linear Regression & \xmark  & 3 Hours & 1.70 & 2.13 & 1.61 & 1.25 & 3.12 & 2.22 & 4.92 & 3.48 \\
        Linear Regression & \xmark  & 4 Hours & 1.66 & 2.22 & 1.61 & \cellcolor[gray]{0.9}\textbf{1.07} & 3.02 & 1.75 & 4.79 & 3.01 \\
        Linear Regression & \xmark & 5 Hours & 1.66 & 2.22 & 1.59 & 1.07 & 3.02 & 1.75 & 4.82 & 3.01 \\
        Linear Regression & \checkmark & 1 Hour & 0.99 & \cellcolor[gray]{0.9}\textbf{0.80} & 1.69 & 1.85 & 2.94 & 3.28 & 4.62 & 4.96 \\
        Linear Regression & \checkmark & 2 Hours & 0.48 & 1.71 & 0.64 & 2.44 & \cellcolor[gray]{0.9}\textbf{2.02} & 3.16 & 3.17 & 3.22 \\
        Linear Regression & \checkmark & 3 Hours & 0.48 & 1.99 & 0.64 & 2.59 & 2.04 & 3.05 & 3.22 & 3.15 \\
        Linear Regression & \checkmark & 4 Hours & 0.46 & 1.90 & 0.63 & 2.63 & 2.05 & 3.12 & 3.24 & 3.19 \\
        Linear Regression & \checkmark & 5 Hours & 0.46 & 1.83 & \cellcolor[gray]{0.9}\textbf{0.62} & 2.75 & 2.04 & 3.24 & 3.24 & 3.19 \\
        XGBoost & \xmark  & 1 Hour & 0.43 & 1.89 & 1.18 & 1.49 & 3.22 & 3.27 & 4.45 & 4.76 \\
        XGBoost & \xmark  & 2 Hours & 0.31 & 2.60 & 3.29 & 1.80 & 4.21 & 3.37 & 4.15 & 2.85 \\
        XGBoost & \xmark  & 3 Hours & 0.94 & 2.77 & 3.29 & 2.17 & 3.93 & 1.28 & 3.50 & 3.04 \\
        XGBoost & \xmark  & 4 Hours & 1.67 & 2.27 & 2.90 & 1.71 & 1.75 & \cellcolor[gray]{0.9}\textbf{1.27} & 4.20 & \cellcolor[gray]{0.9}\textbf{2.68} \\
        XGBoost & \xmark  & 5 Hours & 0.68 & 2.89 & 2.39 & 2.07 & 2.96 & 2.43 & \cellcolor[gray]{0.9}\textbf{2.57} & 6.40 \\
        XGBoost & \checkmark & 1 Hour & 0.86 & 1.27 & 0.88 & 1.33 & 3.14 & 2.61 & 4.93 & 5.38 \\
        XGBoost & \checkmark & 2 Hours & \cellcolor[gray]{0.9}\textbf{0.17} & 2.81 & 0.97 & 2.22 & 3.62 & 2.52 & 4.72 & 4.68 \\
        XGBoost & \checkmark & 3 Hours & 2.55 & 3.66 & 1.05 & 2.71 & 3.71 & 3.58 & 4.84 & 4.89 \\
        XGBoost & \checkmark & 4 Hours & 2.47 & 3.00 & 1.38 & 2.50 & 4.17 & 3.13 & 4.81 & 4.49 \\
        XGBoost & \checkmark & 5 Hours & 2.38 & 2.80 & 1.20 & 2.43 & 3.84 & 2.93 & 4.69 & 4.51 \\
        \bottomrule
    \end{tabular}
\end{table*}

\subsection{Analytics and Alerting System}\label{sec:aas-evaluation}

\subsubsection{Forecasting Model}
To train and evaluate our forecasting models, we utilized data collected from the environmental sensors. Temperature and relative humidity sensors are integrated into ESP32-S3 modules running custom firmware, which provide readings every few minutes. A total of six such modules are installed throughout the poultry house, and the system uses the median of their readings to ensure robustness against sensor noise or potential malfunctions. The dataset consisted of sensor readings from the past 100 days. The first 80 days were used for training, while the remaining 20 days served as the test set to assess model performance.

For each sensor, we computed the hourly average to ensure consistent temporal granularity and to mitigate the effects of missing values. To further enhance robustness, we used the median of all sensor readings at each hour as the representative metric, thereby reducing the influence of noise and outliers caused by potential sensor malfunctions.  

To identify the most effective approach for forecasting temperature and humidity, we evaluated two regression models: Linear Regression~\citep{pedregosa2011scikit} and XGBoost~\citep{chen2016xgboost}. For each algorithm, we tested two variations—one that incorporated the historical profile of the forecasted time and one that did not. Additionally, for both variations, we explored the impact of including look-back window of the target variable up to 5 hours, resulting in 10 configurations per model.

The historical profile was constructed by averaging the temperature and humidity values recorded at the same hour of the day over the previous 3, 7, and 14 days. This allowed the model to capture recurring daily patterns and seasonal trends relevant to the specific forecast horizon. Examples of historical profiles for temperature and humidity are shown in Figures~\ref{fig:Temp_Profile} and \ref{fig:Hum_Profile} respectively.

The performance of each model configuration for forecasting temperature and humidity across multiple time horizons is presented in Table \ref{tab:short_term}. Across all configurations, models that included the historical profile consistently outperformed those that did not. However, the optimal combination of model type and look-back window varied depending on the forecast horizon and the specific target variable. For short-term forecasts up to 3 hours, the Linear Regression models provided sufficient accuracy for our purposes. Given the computational constraints of our system, which operates on a Raspberry Pi 5 with limited processing capacity, we opted to use the Linear Regression model to maintain both efficiency and reliability.

\subsubsection{Early Warning System Thresholds}
To develop an effective Early Warning System, it was essential to establish appropriate thresholds for each metric monitored by our modules. For environmental sensor data, threshold values were determined based on relevant literature and expert opinions. In contrast, identifying optimal thresholds for audio and video metrics required an experimental approach to ensure accurate and reliable alert generation.

\paragraph{Environmental Metrics Thresholds:}
According to \cite{efsa2023}, the ideal temperature range for poultry lies between 20°C and 25°C, while heat stress begins to affect production at temperatures above 30°C. Taking this into consideration, along with expert insights and the specific climatic conditions in Cyprus, we established an upper threshold of 35°C for triggering heat stress alerts, and a lower threshold of 18°C for cold stress alerts.

Similarly, optimal relative humidity levels for poultry production fall between 40\% and 60\%, as values outside this range can negatively impact productivity and increase the risk of disease within the flock. Based on this, we set the lower and upper thresholds for humidity alerts at 40\% and 60\%, respectively.

\begin{figure*}[ht]
    \centering
    \begin{subfigure}[b]{\linewidth}
        \centering
        \includegraphics[width=\linewidth]{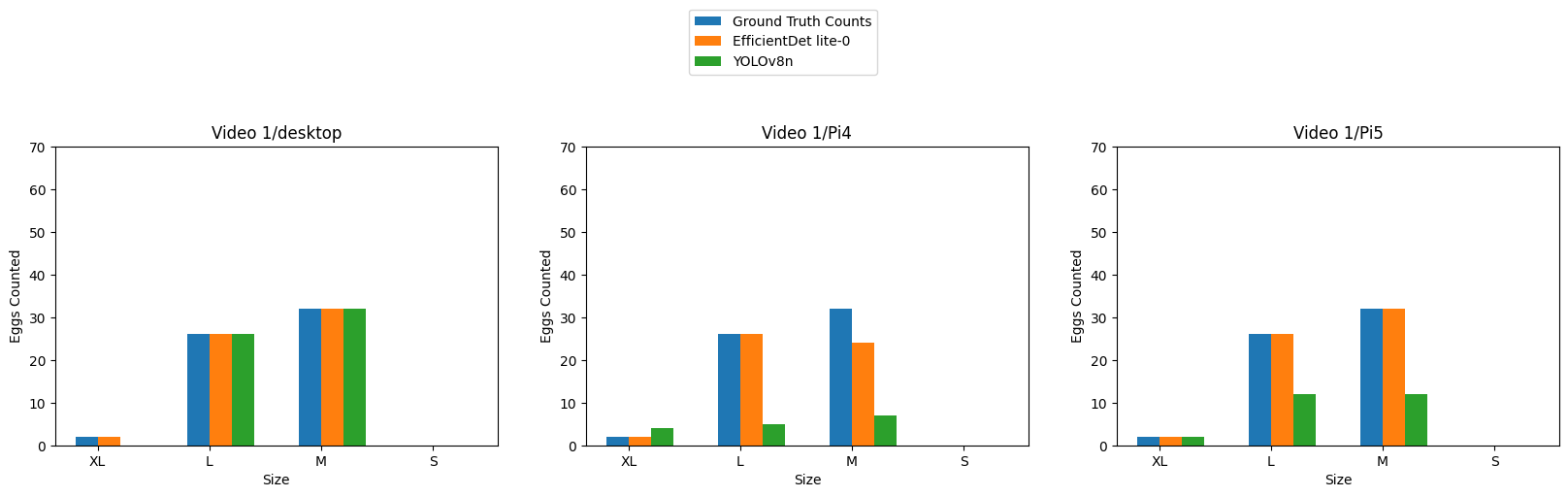}
        \label{fig:subfig1}
    \end{subfigure}
    
    \begin{subfigure}[b]{\linewidth}
        \centering
        \includegraphics[width=\linewidth]{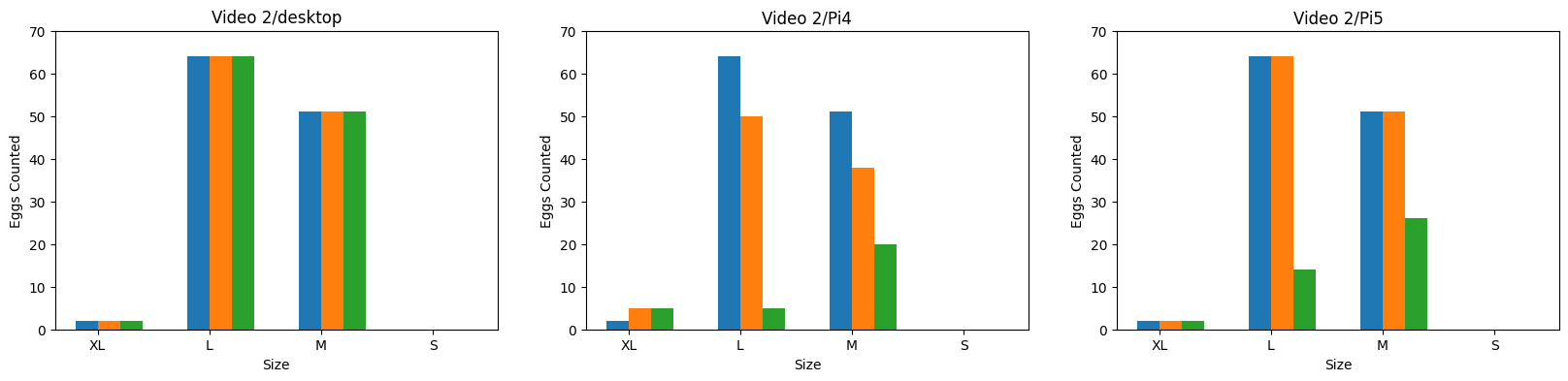}
        \label{fig:subfig2}
    \end{subfigure}
    
    \begin{subfigure}[b]{\linewidth}
        \centering
        \includegraphics[width=\linewidth]{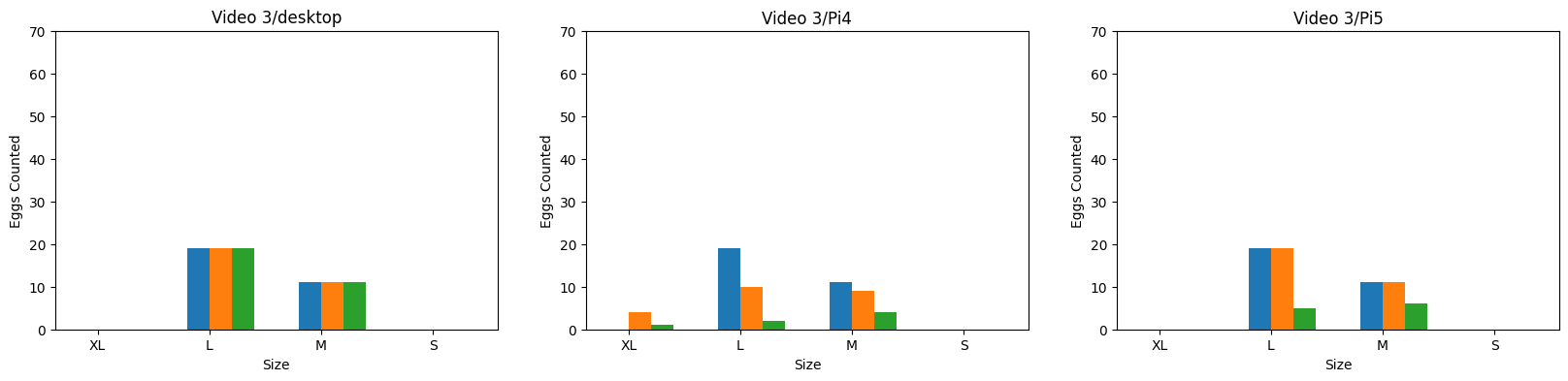}
        \label{fig:subfig3}
    \end{subfigure}
    
    \caption{Produced results for all three devices and models over three different test videos. Each row represents processing of a different video. Figure taken from~\citep{hadjisavvas2024eggcounting}.}
    \label{fig:results}
\end{figure*}

\paragraph{Audio and Video Metrics Thresholds:}
Poultry behavior varies significantly throughout the day, with increased movement typically observed during feeding or egg collection times, and reduced activity and noise levels during nighttime rest periods. Consequently, applying a single, static threshold to the audio and video metrics across the entire day proved ineffective.

To address this, we calculated weighted average values for each minute of the day. For each time point, the value incorporates data from the surrounding 60-minute window, both before and after the target minute, giving greater weight to values closer to the target time. The upper threshold was defined as the 95th percentile of these weighted averages, while the lower threshold was set at the 25th percentile. To determine the appropriate thresholds, we conducted a series of tests using various threshold values on a two-month test set of metric data. The primary objective was to ensure that the system would notify farmers only in response to meaningful or abnormal events—while avoiding alerts during routine activities such as regular feeding or egg collection. Stricter thresholds tended to overlook meaningful events, while more lenient thresholds resulted in an excessive number of daily alerts, many of which were false positives.

\begin{figure}[ht]
    \centering
    \includegraphics[width=1\linewidth]{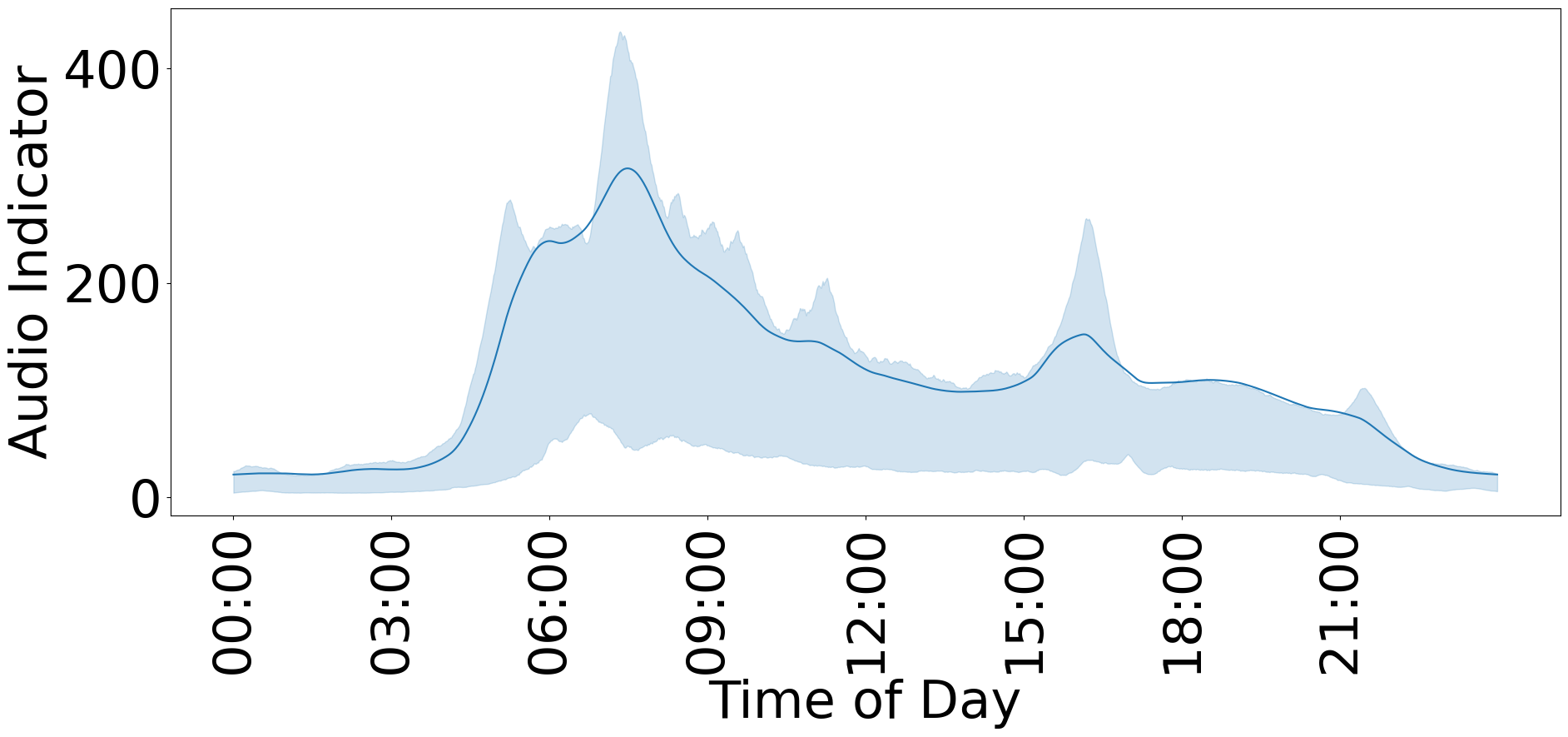}
    \caption{The Profile of the Audio Indicator Averages during the Day.  The solid line represents the Weighted Average Value and the shaded region represents the values within the 25th and 95th percentile.}
    \label{fig:Audio_Profile}
\end{figure}

\begin{figure}[ht]
    \centering
    \includegraphics[width=1\linewidth]{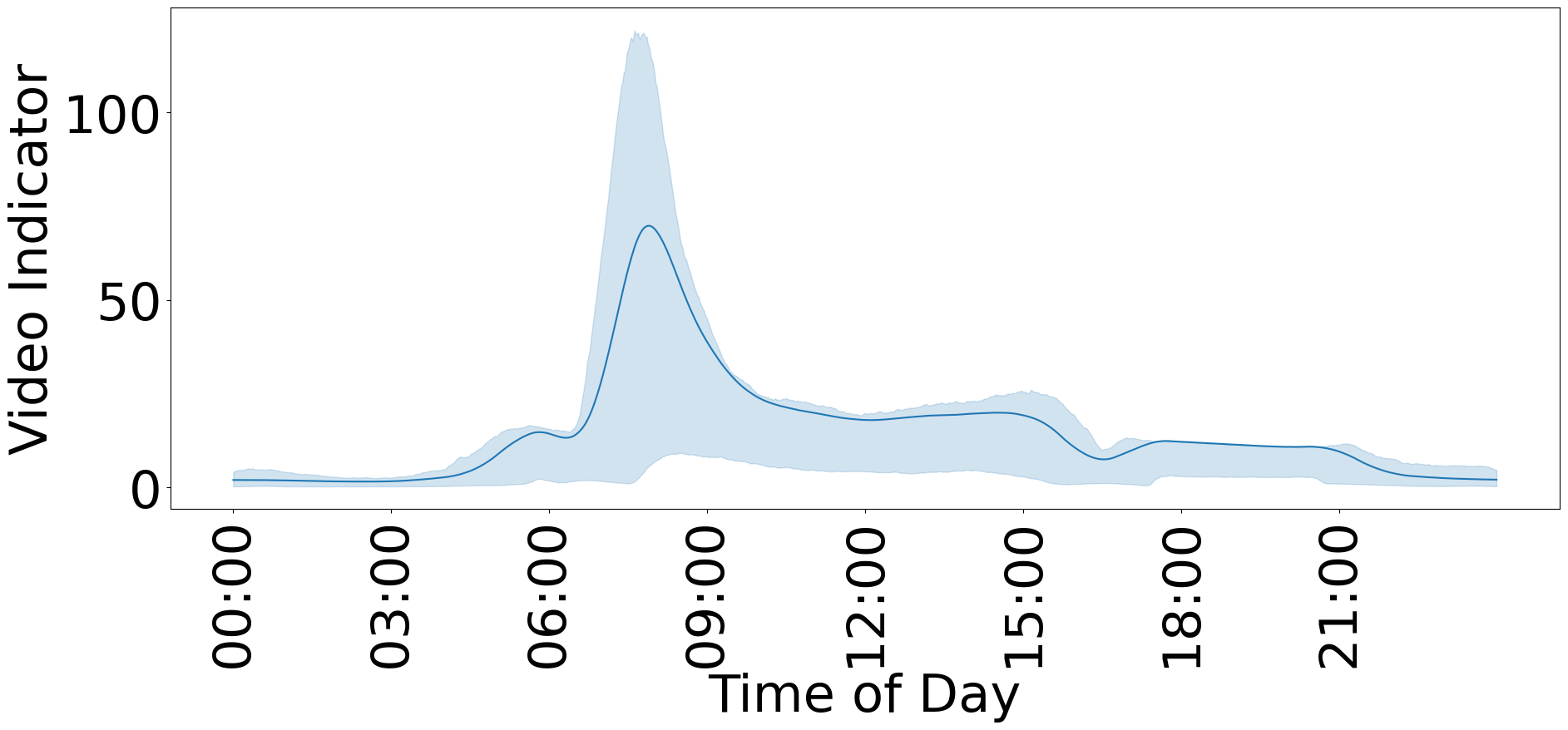}
    \caption{The Profile of the Video Indicator Averages during the Day. The solid line represents the Weighted Average Value and the shaded region represents the values within the 25th and 95th percentile.}
    \label{fig:Video_Profile}
\end{figure}

This approach results in dynamic, time-specific thresholds that more accurately capture the expected behavioral patterns of the flock based on the time of day. The profiles along with the areas within the thresholds for both metrics are shown in Figure \ref{fig:Audio_Profile} and Figure \ref{fig:Video_Profile}.

\begin{table*}[ht]
    \centering
    \caption{The 40 features considered for the Egg Production Forecasting model}
    \label{tab:features_productivity}
    \begin{tabular}{lll | lll}
        \toprule 
        \textbf{Feature} & \textbf{Description} & \textbf{Used} &
        \textbf{Feature} & \textbf{Description} & \textbf{Used} \\
        \midrule
        f1  & Avg temp (prev day) & \checkmark &
        f21 & Video during feed (past 3 days) & \xmark \\
        f2  & Max temp (prev day) & \xmark &
        f22 & Audio at night (past 7 days) & \xmark \\
        f3  & Min temp (prev day) & \xmark &
        f23 & Video at night (past 7 days) & \xmark \\
        f4  & Avg humidity (prev day) & \checkmark &
        f24 & Audio at night (past 3 days) & \xmark \\
        f5  & Max humidity (prev day) & \xmark &
        f25 & Video at night (past 3 days) & \xmark \\
        f6  & Min humidity (prev day) & \xmark &
        f26 & Audio at rest (past 7 days) & \xmark \\
        f7  & Chicken age (weeks) & \checkmark &
        f27 & Video at rest (past 7 days) & \xmark \\
        f8  & Avg audio (past 15 days) & \checkmark &
        f28 & Audio at rest (past 3 days) & \checkmark \\
        f9  & Avg video (past 15 days) & \xmark &
        f29 & Video at rest (past 3 days) & \xmark \\
        f10 & Avg audio (past 7 days) & \checkmark &
        f30 & Top-5 audio feed (past 3 days) & \checkmark \\
        f11 & Avg video (past 7 days) & \xmark &
        f31 & Top-5 video feed (past 3 days) & \checkmark \\
        f12 & Avg audio (past 3 days) & \xmark &
        f32 & Low-5 audio feed (past 3 days) & \checkmark \\
        f13 & Avg video (past 3 days) & \xmark &
        f33 & Low-5 video feed (past 3 days) & \checkmark \\
        f14 & Egg prod avg (past 7 days) & \checkmark &
        f34 & Top-5 audio night (past 3 days) & \xmark \\
        f15 & Egg prod avg (past 3 days) & \checkmark &
        f35 & Top-5 video night (past 3 days) & \xmark \\
        f16 & Dead chickens avg (past 7 days) & \xmark &
        f36 & Top-5 audio rest (past 3 days) & \xmark \\
        f17 & Dead chickens avg (past 3 days) & \checkmark &
        f37 & Top-5 video rest (past 3 days) & \xmark \\
        f18 & Audio during feed (past 7 days) & \xmark &
        f38 & Low-5 audio rest (past 3 days) & \xmark \\
        f19 & Video during feed (past 7 days) & \checkmark &
        f39 & Low-5 video rest (past 3 days) & \xmark \\
        f20 & Audio during feed (past 3 days) & \checkmark &
        f40 & Number of chickens & \xmark \\
        \bottomrule
    \end{tabular}
\end{table*}

\subsection{Real-Time Egg Counting}

We evaluated on three hardware setups, i.e., an Intel i7/RTX 3070 desktop, a Raspberry Pi 4, and a Raspberry Pi 5, using two detectors (EfficientDet-Lite0 and YOLOv8n). Models were trained on 2,226 annotated egg images (2,065 train / 161 val), then tested on three videos of the egg grading and counting procedure in which 30-110 eggs were classified into four commercial weight grades (small, medium, large, extra large eggs).

On the desktop configuration, EfficientDet-Lite0 achieved perfect counts in all three videos; YOLOv8n missed only two extra-large eggs in one session. Raspberry Pi 4 performance lagged: EfficientDet-Lite0 ran at ~0.159 s/frame, YOLOv8n at ~0.515 s/frame, leading to systematic under-counts as the machine outpaced inference. In contrast, on Raspberry Pi 5, EfficientDet-Lite0 ran at ~0.046 s/frame and matched ground-truth counts on all videos (100\% accuracy). YOLOv8n on Pi 5 managed only partial accuracy (e.g., missing multiple medium and large eggs), running at ~0.256 s/frame. See Figure~\ref{fig:results} for a visualisation of the results.

These results demonstrate that a calibrated, optimized lighter-weight detector, running on a low cost edge device, can deliver desktop-level egg-counting accuracy, enabling practical, field-deployable automation without GPU acceleration. For more details on the methodology and results, we refer the reader to \citep{hadjisavvas2024eggcounting}.

\subsection{Production and Profitability Forecasting System}\label{sec:ppf-evaluation}

To identify the most effective model and feature set for our egg production forecasting system, we experimented with two regression algorithms: Linear Regression and XGBoost. Various feature configurations were also tested to assess the contribution of different data sources. All the data were collected daily and were used to predict the average productivity for the next 10 days.

Initially, for each model, we evaluated a configuration that relied solely on production data provided by the farmers—specifically, flock age, egg production history, and mortality rates. This configuration was tested using data, covering a period of three years, during which no sensor data was available (Dataset A).

Subsequently, we evaluated a configuration that incorporated environmental sensor data, as described in Section~\ref{sec:aas-description}. This configuration included all the features from the initial setup, along with daily minimum, maximum, and average temperature readings. It was tested on data collected from a one year period (Dataset B).

The final configuration extended the previous one by also including audio and video metrics from the models described in Section~\ref{sec:audiovisual}. This configuration was tested on data collected from a 4 month period (Dataset C). We first considered 40 hand-engineered features from all sources. After experimenting with different subsets of these features, we identified those that contribute most to prediction performance and those were used for Dataset C. The full list of the features considered and used can be found in Table~\ref{tab:features_productivity}. For consistency, the earlier configurations (Production data only, production + environmental data) were also evaluated on this same dataset, further confirming that performance on the test set improved as additional data sources were integrated.  

\begin{figure}[ht]
  \centering
  \includegraphics[width=1\linewidth]{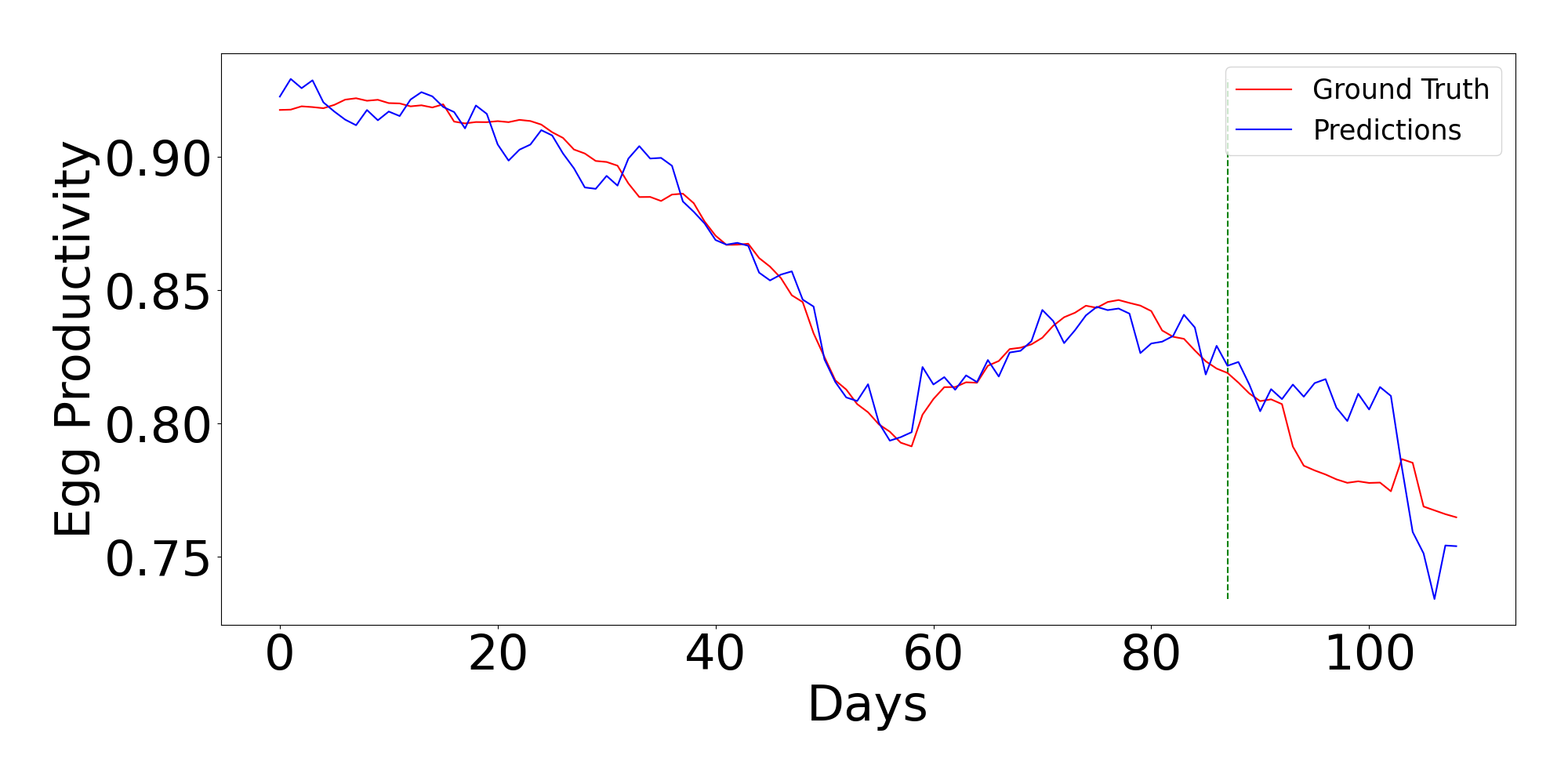}
  \caption{The predictions of the final egg productivity model. The predictions follow the general data tendency while the error is less than 2\% of the average daily egg production. The dashed line separates the training set from the test set.}
  \label{fig:EggProductionLR}
\end{figure}

For evaluation, each dataset was split chronologically, using the first 80\% of records for training and the remaining 20\% for testing. The Mean Absolute Error (MAE) on the test set for egg productivity forecasts is reported in Table~\ref{tab:productivity}. As shown, model performance improved as features from additional sources were introduced. Despite the reduced amount of training data, the configurations that included all available data sources outperformed the others. Moreover, the Linear Regression model outperformed XGBoost on the last configuration of the models. Based on these results, we selected Linear Regression for integration into our system, using Dataset C. The performance of the model is shown in Figure~\ref{fig:EggProductionLR}.

\begin{table*}[h]
    \centering
    \caption{Comparison between the different model configurations using the Test Set MAE on Productivity}
    \label{tab:productivity}
    \begin{tabular}{llcccccc c}
        \toprule
         \textbf{Data}& \textbf{Model} & \multicolumn{2}{c}{\textbf{Production Data}} & \multicolumn{2}{c}{\textbf{Sensor Data}} & \multicolumn{2}{c}{\textbf{Audio/Video Data}} & \textbf{Test Set MAE} \\
        & & Available & Used & Available & Used & Available & Used & \\
        \midrule
        Dataset A & Linear Regression & \checkmark & \checkmark & \xmark & \xmark & \xmark & \xmark & 0.08 \\
        Dataset A & XGBoost           &  \checkmark & \checkmark & \xmark & \xmark & \xmark & \xmark & 0.13 \\
        Dataset B & Linear Regression &  \checkmark & \checkmark & \checkmark & \checkmark & \xmark & \xmark & 0.07 \\
        Dataset B &  XGBoost           &  \checkmark & \checkmark & \checkmark & \checkmark & \xmark & \xmark & 0.04 \\
        Dataset C & Linear Regression &  \checkmark & \checkmark & \checkmark & \xmark & \checkmark & \xmark & 0.05 \\
        Dataset C & XGBoost           & \checkmark & \checkmark & \checkmark & \xmark & \checkmark & \xmark & 0.03 \\
        Dataset C & Linear Regression & \checkmark & \checkmark & \checkmark & \checkmark & \checkmark & \xmark & 0.04 \\
        Dataset C & XGBoost           & \checkmark & \checkmark & \checkmark & \checkmark & \checkmark & \xmark & 0.03 \\
        Dataset C & Linear Regression & \checkmark & \checkmark & \checkmark & \checkmark & \checkmark & \checkmark & \cellcolor[gray]{0.9}\textbf{0.01} \\
         Dataset C &  XGBoost           & \checkmark & \checkmark & \checkmark & \checkmark & \checkmark & \checkmark & 0.03 \\
        \bottomrule
    \end{tabular}
\end{table*}

\subsection{Recommendation Module}

Recommendations related to Environmental Sensors and the Audio-Video Analytics are generated based on the thresholds outlined in Section~\ref{sec:aas-evaluation}. For weather-based recommendations, the system considers four key factors:

\begin{itemize} 
\item \textbf{Temperature}: An alert is triggered when the forecasted temperature is above 35°C (heat) or below 10°C (cold). 
\item \textbf{Wind}: An alert is triggered when the forecasted wind speed is above 29 km/h. 
\item \textbf{Cloud Coverage}: An alert is triggered when the forecasted cloud coverage is above 70\%. 
\item \textbf{Rain Chance}: An alert is triggered when the forecasted chance of precipitation is above 50\%. 
\end{itemize}

\begin{figure}[ht]
\centering
\scriptsize

\tikzset{
    box/.style={
        rectangle, draw=black, rounded corners, thick, fill=blue!5,
        text width=0.46\columnwidth, align=left, inner sep=5pt
    },
    arrow/.style={
        thick, -{Latex[length=2mm]}
    }
}

\begin{tikzpicture}[node distance=0.8cm and 0.3cm]

\node[box, label=above:{\textbf{Case 1}}] (cond1) {
    \textbf{Conditions:}
    \begin{itemize}[leftmargin=*, itemsep=2pt]
        \item Area Temperature: 15°C – 27°C
        \item Max cloud coverage: 98\%
        \item Max rain chance: 0\%
        \item Max wind speed: 22 km/h
        \item Farm temperature: 26°C
        \item Farm humidity: 36\%
        \item Audio indicator: Low
        \item Video indicator: Low
        \item Predicted productivity: 77\%
    \end{itemize}
};

\node[box, right=of cond1] (rec1) {
    \textbf{System's Recommendations:}
    \begin{itemize}[leftmargin=*, itemsep=2pt]
        \item Cloudy weather is expected in the next few days. You may want to open the poultry house's lights.
        \item Humidity is expected to be very low. Turn on the water sprayers.
        \item The flock is less active than expected. Check food and water intake.
    \end{itemize}
};
\draw[arrow] (cond1) -- (rec1);

\node[box, below=of cond1, label=above:{\textbf{Case 2}}] (cond2) {
    \textbf{Conditions:}
    \begin{itemize}[leftmargin=*, itemsep=2pt]
        \item Area Temperature: 20°C – 36°C
        \item Max cloud coverage: 30\%
        \item Max rain chance: 5\%
        \item Max wind speed: 15 km/h
        \item Farm temperature: 38°C
        \item Farm humidity: 65\%
        \item Audio indicator: High
        \item Video indicator: High
        \item Predicted productivity: 65\%
    \end{itemize}
};

\node[box, right=of cond2] (rec2) {
    \textbf{System's Recommendations:}
    \begin{itemize}[leftmargin=*, itemsep=2pt]
        \item Warning! The egg productivity is predicted to fall under 70\% in the next few days.
        \item Extreme heat is expected in the next few days. You may want to check your fans and water sprayers.
        \item The temperature in the farm is expected to be very high in the next hour. Turn on the fans.
        \item The humidity in the farm is expected to be very high in the next hour. You need to add bedding in the poultry house.
        \item There is too much noise or movement in the barn. This could affect the egg production.
    \end{itemize}
};
\draw[arrow] (cond2) -- (rec2);

\node[box, below=of cond2, label=above:{\textbf{Case 3}}] (cond3) {
    \textbf{Conditions:}
    \begin{itemize}[leftmargin=*, itemsep=2pt]
        \item Area Temperature: 5°C – 20°C
        \item Max cloud coverage: 95\%
        \item Max rain chance: 85\%
        \item Max wind speed: 30 km/h
        \item Farm temperature: 10°C
        \item Farm humidity: 45\%
        \item Audio indicator: No alert
        \item Video indicator: No alert
        \item Predicted productivity: 72\%
    \end{itemize}
};

\node[box, right=of cond3] (rec3) {
    \textbf{System's Recommendations:}
    \begin{itemize}[leftmargin=*, itemsep=2pt]
        \item Wind is expected in the next few days. You may want to close the barn's windows.
        \item Extreme cold is expected in the next few days. You may want to check your heaters.
        \item The temperature in the farm is expected to be very low in the next hour. Turn on the heaters.
    \end{itemize}
};
\draw[arrow] (cond3) -- (rec3);

\end{tikzpicture}

\caption{Conditions and corresponding system recommendations for three distinct cases.}
\label{fig:recommendation_system}
\end{figure}
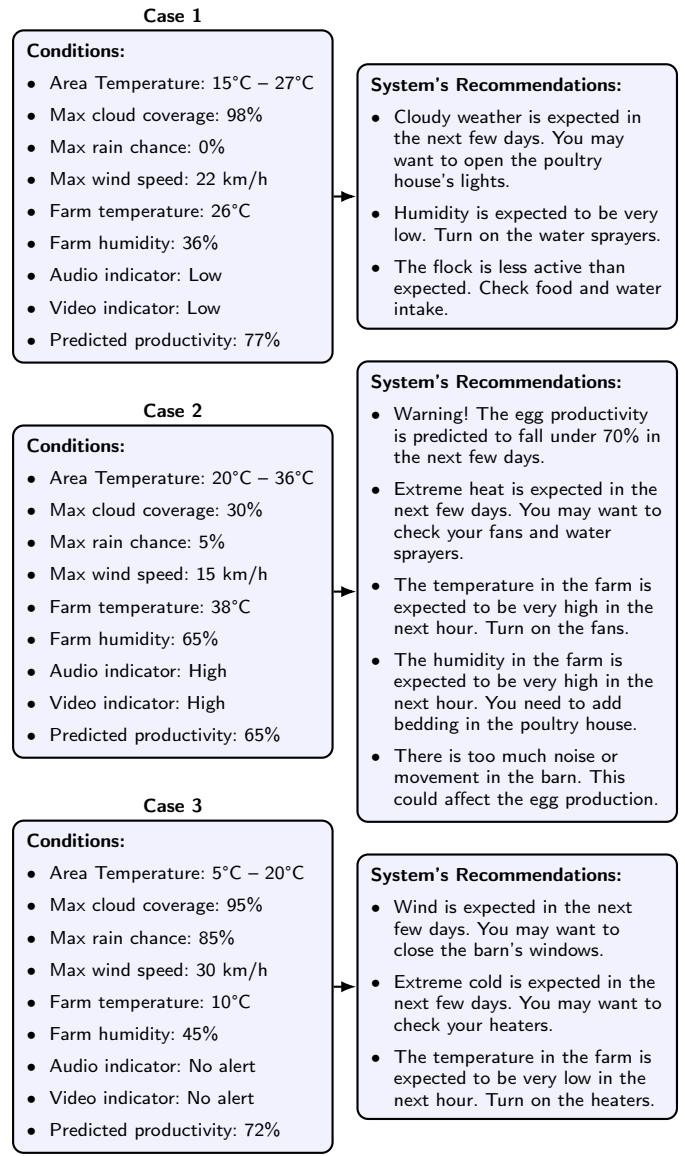

Different test cases of the system in different conditions are shown in Figure \ref{fig:recommendation_system}.

\section{Discussion}\label{sec-discussion}

\subsection{Lessons Learned}
Integrating multiple AI‐driven modules into a cohesive system yielded several key insights. First, adopting a modular architecture allowed us to develop, test, and refine each component independently while maintaining clear data interfaces—a practice that greatly accelerated iteration and debugging. Second, combining heterogeneous sensing modalities (video, audio, environmental) proved more powerful than relying on any single data stream: where motion detection might miss subtle signs of stress, audio anomalies filled the gap, and vice versa. Third, deploying inference at the edge (on Raspberry Pi devices) validated that real‐time welfare monitoring and egg‐counting are feasible without constant cloud connectivity, thereby reducing latency and preserving data privacy. Overall, these lessons underscore the value of multi‐modal, distributed intelligence in precision livestock farming.

\subsection{Challenges and Limitations}
Despite the advancement presented in this work, further investigation is necessary to tackle challenges that remain. 
Audio‐based models encounter variability in ambient noise: feeders, ventilation fans, and equipment cycles can generate false positives, necessitating careful feature engineering and smoothing strategies. Visual monitoring faces its own constraints: changes in lighting, dust accumulation on lenses, and occlusions during high bird density can degrade motion metrics. Finally, model performance depends on the quality and representativeness of training data; rare events (e.g., acute disease outbreaks) remain difficult to capture without extensive, labeled datasets.

\subsection{Scalability and Generalizability}
PoultryFI’s core modules are designed to scale and adapt to different operational scales and species. For larger operations requiring additional cameras, the Camera Placement Optimizer can accommodate any number of devices; however, processing more video and audio streams will exceed the capacity of a single Raspberry Pi 5. In such cases, the AVMM can be extended by deploying multiple synchronized Processing Engines (e.g., additional Pi 5 units) or migrating inference to more capable edge platforms (for example, NVIDIA Jetson Orin). Beyond poultry layers, the same principles, i.e., optimal sensor deployment, multi‐modal welfare monitoring, short‐ and long-term predictive analytics, and context-aware recommendations, can be applied to free-range flocks or other livestock such as sheep and cattle.
Nevertheless, the system’s generalizability across different farm environments and conditions remains an open question. Further investigation is needed to evaluate its adaptability to varying setups, management practices, and potentially other animal species.

\subsection{Future Directions}
Looking forward, several avenues promise to extend PoultryFI’s impact. On the research front, deeper integration, such as co-training forecasting models with welfare indicators or exploring reinforcement-learning controllers for automated climate and feeding adjustments, could further improve responsiveness. Furthermore, the existing system should be evaluated in a wider range of farms featuring diverse weather conditions, sizes, capacities, and chicken breeds to better assess its adaptability. In addition, the impact of the Recommendation Module on both productivity and animal welfare should be studied once it is fully integrated into the farm's daily operations for a longer period of time.

Commercially, PoultryFI’s modular design supports phased adoption: producers might begin with welfare monitoring and alerts before deploying forecasting or recommendation features. Integrations with farm-management software and mobile dashboards will streamline user workflows and facilitate broader uptake.

\section{Conclusion}\label{sec-conclusion}

In this paper, we introduced the Poultry Farm Intelligence (PoultryFI) system, a low-cost, fully integrated platform comprising six novel modules: the Camera Placement Optimizer for automated, high‐coverage surveillance layouts; the Audio-Visual Monitoring Module (AVMM) for non-invasive welfare indicators derived from motion, sound, and feeding data; the Analytics \& Alerting Module (AAM) for real-time and forecasted anomaly detection across behavioral and environmental streams; the Real-Time Egg Counting Module for accurate, edge-based egg counts; the Production \& Profitability Forecasting Module (PPFM) for short- to medium-term projections of egg output and cost-per-egg; and the Recommendation Module for prescriptive, context-aware management guidance. Together, these modules demonstrate how multi-modal sensing, AI-driven analytics, and edge-computing can be combined to advance both operational efficiency and animal welfare in poultry farming.

\section*{Acknowledgments}
The work was funded by the European Union
Recovery and Resilience Facility of the NextGenerationEU instrument, through the Research and Innovation Foundation (CODEVELOP-ICT-HEALTH/0322/0061), as well as the European Union’s Horizon 2020 Research and Innovation Programme under Grant Agreement No. 739578, and the Government of the Republic of Cyprus through the Deputy Ministry of Research, Innovation and Digital Policy.

\bibliographystyle{cas-model2-names}

\bibliography{refs}

\end{document}